\documentclass[twoside]{article}

\usepackage[bindingoffset=0cm,a4paper,centering,textheight=200mm,textwidth=120mm,includefoot,includehead]{geometry}
\usepackage[all]{xy}
\usepackage{tikz}
\usepackage{float}
\usepackage{euscript}
\usepackage{natbib}
\usepackage{epstopdf}
\usepackage{chapterbib}
\usepackage{textcomp}
\usepackage{times}
\usepackage{amssymb}
\usepackage{amsfonts}
\usepackage{latexsym}
\usepackage{url}
\usepackage{epsfig}
\usepackage{amsmath}
\usepackage{amsfonts}
\usepackage{graphics}
\usepackage{graphicx}
\usepackage{makeidx}
\usepackage{psfrag}
\usepackage{array}
\usepackage{tabularx}
\usepackage{xspace}
\usepackage{wrapfig}

\usepackage{times}
\usepackage[scaled]{helvet}
\usepackage{courier}

\usepackage{color}
\usepackage{booktabs}
\usepackage{subfig}

\usepackage{multirow}

\usepackage{ifpdf}

\usepackage{paralist}
\usepackage{enumerate}

\newcommand{\ignore}[1]{}

\makeatother

\begin{document}

\newcommand{\MDD}{McDermott and Doyle\xspace}
\newcommand{\onKon}{$\stackrel{Kon}{\longrightarrow}$}
\newcommand{\Kon}{\mathit{Kon}}
\newcommand{\Theo}{\mathit{Th}}
\newcommand{\st}{\;|\;}
\newcommand{\vph}{\varphi}

\newcommand{\la}{\leftarrow}

\newcommand{\wis}[1]{\ensuremath{#1}} 
\newcommand{\curly}[1]{\mathcal{#1}}

\newcommand{\cent}[1]{\begin{center}#1\end{center}} 
\newcommand{\verz}[1]{\{#1\}}

\newcommand{\itz}{\begin{itemize}\item } 
\newcommand{\eitz}{\end{itemize}} 
\newcommand{\Tr}{{\mbox{\bf t}}}
\newcommand{\Fa}{{\mbox{\bf f}}}
\newcommand{\Un}{{\mbox{\bf u}}}
\newcommand{\In}{{\mbox{\bf i}}}

\newcommand{\nmdash}{|\!\!\!\sim}

\newcommand{\AIKA}{{\em All I Know} Assumption\xspace}

\newcommand{\D}{{\curly{D}}}

\newcommand{\mim}{\rightarrow}  
\newcommand{\ra}{\rightarrow}
\newcommand{\Ra}{\Rightarrow}
\newcommand{\leqp}{\leq_p}

\newcommand{\Worlds}{\curly{W}}

\newcommand{\tbs}{{\mathcal{B}}}
\newcommand{\bs}{B}

\newcommand{\HH}[2]{|#2|{#1}}
\newcommand{\tf}[2]{{|#1|}^{#2}}
\newcommand{\stf}[2]{|{#1}|^{#2}_{sv}}
\newcommand{\ktf}[2]{|{#1}|^{#2}_{K}}

\newcommand{\struct}[1]{\langle#1\rangle}
\newcommand{\default}[2]{\begin{array}{c}#1\\\hline #2\end{array}}

\newtheorem{example}{Example}
\newtheorem{definition}{Definition}
\newtheorem{proposition}{Proposition}

\setcounter{equation}{0}
\setcounter{footnote}{0}
\setcounter{definition}{0}
\setcounter{section}{0}
\markboth{M.~Denecker, V.~W.~Marek and M.~Truszczy{\'n}ski}{Reiter's Default Logic
Is a Logic of Autoepistemic Reasoning}
\thispagestyle{empty}

\ \\
\ \\
{\LARGE\bf Reiter's Default Logic Is a Logic of Autoepistemic
Reasoning And a Good One, Too }

\ \\
\ \\
\ \\
\ \\
{\large\textbf{Marc Denecker}}\\
Department of Computer Science\\
K.U. Leuven\\
Celestijnenlaan 200A\\
B-3001 Heverlee, Belgium

\medskip
\noindent
{\large\textbf{Victor W. Marek}}\\
{\large\textbf{Miros{\l}aw Truszczy{\'n}ski}}\\
Department of Computer Science\\
University of Kentucky\\
Lexington, KY 40506-0633, USA

\medskip
\ \\
\noindent
{\textbf{Abstract:}
A fact apparently not observed earlier in the literature of nonmonotonic 
reasoning is that Reiter, in his default logic paper, did not directly 
formalize \emph{informal} defaults. Instead, he translated a default 
into a certain natural language proposition and provided a formalization
of the latter. A few years later, Moore noted that propositions like the
one used by Reiter are fundamentally different than defaults and exhibit
a certain \emph{autoepistemic} nature. Thus, Reiter had developed his 
default logic as a formalization of autoepistemic propositions rather 
than of defaults.

The first goal of this paper is to show that some problems of Reiter's
default logic as a formal way to reason about informal defaults are
directly attributable to the autoepistemic nature of default logic and 
to the mismatch between informal defaults and the Reiter's formal 
defaults, the latter being a formal expression of the autoepistemic 
propositions Reiter used as a representation of informal defaults.

The second goal of our paper is to compare the work of Reiter and 
Moore. While each of them attempted to formalize autoepistemic 
propositions, the modes of reasoning in their respective logics were 
different. We revisit Moore's and Reiter's intuitions and present them 
from the perspective of {\em autotheoremhood}, where theories can 
include propositions referring to the theory's own theorems. We then 
discuss the formalization of this perspective in the logics of Moore 
and Reiter, respectively, using the unifying semantic framework for 
default and autoepistemic logics that we developed earlier. We argue 
that Reiter's default logic is a better formalization of Moore's 
intuitions about autoepistemic propositions than Moore's own 
autoepistemic logic.
}


\section{Introduction}
\label{Intro}

In this volume we celebrate the publication in 1980 of the special
issue of the Artificial Intelligence Journal on Nonmonotonic Reasoning
that included three seminal papers: \emph{Logic for Default
  Reasoning} by \citet{re80}, \emph{Nonmonotonic Logic I}
by \citet{McDermott80}, and \emph{Circumscription --- a form of
  nonmonotonic reasoning} by \citet{McCarthy80}. While the roots of the
subject go earlier in time, these papers are universally viewed as the
main catalysts for the emergence of nonmonotonic reasoning as a
distinct field of study. Soon after the papers were published,
nonmonotonic reasoning attracted widespread attention of researchers
in the area of artificial intelligence, and established itself firmly
as an integral sub-area of knowledge representation. Over the years,
the appeal of nonmonotonic reasoning went far beyond artificial
intelligence, as many of its research challenges raised
fundamental questions to philosophers and mathematical logicians,
and stirred substantial interest in those communities.

The groundbreaking paper by \citet{McCarthy69} about ten
years before had captured the growing concern with the logical
representation of {\em common sense knowledge}. Attention focused on
the representation of {\em defaults}, propositions that are true for
most objects\ignore{with few exceptions}, that commonly assume the form
\emph{``most $A$'s are $B$'s.''}\footnote{In this paper, we interpret
  the term ``default'' as an informal statement \emph{``most 
$A$'s are $B$'s''} \citep{re80}.  The term is sometimes interpreted
  more broadly to capture {\em communication conventions}, {\em frame 
  axioms} in temporal reasoning, or statements such as ``{\em 
  normally} or {\em typically, A's are B's}''.}  
Defaults arise in all applications
involving common sense reasoning and require specially tailored forms
of reasoning.
For instance, a default \emph{``most $A$'s are $B$'s''}
under suitable circumstances should enable one to infer from the premise 
\emph{``$x$ is an $A$''} that \emph{``$x$ is a $B$.''} This inference 
is {\em defeasible}. Its consequent ``$x$ is a $B$'' may be false 
even if its premise ``$x$ is an $A$'' is true. It may have to be withdrawn
when new information is obtained. Providing a 
general, formal, domain independent and
elaboration tolerant representation of defaults and an account of
what inferences can be {\em rationally} drawn from them was
the artificial intelligence challenge of the time.

The logics proposed by McCarthy, Reiter, and \MDD were developed in an
attempt to formalize reasoning where defaults are present. They
went about it in different ways, however. McCarthy's \emph{circumscription} 
extended a set of 
first-order sentences with a second-order axiom asserting {\em minimality} 
of certain predicates, typically of {\em
abnormality predicates} that capture the exceptions to defaults. This 
reflected the assumption that the world deviates as little 
as possible from the ``normal'' state. Circumscription has played a 
prominent role in nonmonotonic reasoning. In particular, it has been a 
precursor to preference logics \citep{sho87} that provided further important 
insights into reasoning about defaults. 

\citet{re80} and \citet{McDermott80}, on the
other hand, focused on the inference pattern \emph{``most $A$'s are $B$'s.''}
In Reiter's words \citep[p. 82]{re80}:
\begin{quote}
`We take it [that is, the default ``Most birds can fly'' --- DMT
to mean something like ``If an $x$ is a bird,
  then in the absence of any information to the contrary, infer that $x$
  can fly.''' 
\end{quote}
Thus, Reiter (and also \MDD) quite literally equated a default 
\emph{``most $A$'s are $B$'s''} with an inference rule that involves,
besides the premise \emph{``$x$ is an $A$''}, an additional premise 
``there is no information to the contrary'' or, more specifically,
\emph{``there is no information indicating that ``$x$ is not a $B$}.''
The role of this latter premise, a \emph{consistency} condition, is 
to ensure 
the rationality of applying the default.
In logic, inference rules are meta-logical objects that 
are not expressed in a logical language. Reiter, \MDD sought to develop 
a logic in which such meta-logical inference rules could be stated 
\emph{in} the logic itself. They equipped their logics with a suitable 
\emph{modal operator} (in the case of Reiter, embedded within ``his'' 
default expression) to be able to express the \emph{consistency} condition
and, in place of a default \emph{``most $A$'s are $B$'s''}, they used the
statement \emph{``if $x$ is an $A$ and if it is consistent (with the 
available information) to assume that $x$ is a $B$, then $x$ is a $B$.''} 
We will call this latter statement the \emph{Reiter-McDermott-Doyle} 
(RMD, for short) proposition associated with the default.

\citet{mo85} was one of the first, if not the first, who
realized that defaults and their RMD propositions
are of a different nature. This is how \citet[p. 76]{mo85} 
formulated RMD propositions
in terms of theoremhood and non-theoremhood:
\begin{quote} `[In the approaches of McDermott and Doyle, and of
  Reiter --- DMT]
the inference that birds can fly is handled by having, in
  effect, a rule that says that, for any X, ``X can fly'' is a theorem
  if ``X is a bird'' is a theorem and ``X cannot fly'' is {\em not} a
  theorem.'
\end{quote}
Moore then contended that RMD propositions are \emph{autoepistemic} 
statements, that is, introspective statements referring to the reasoner's 
own belief or the theory's own theorems. He pointed out fundamental 
differences between the nature of default propositions and autoepistemic 
ones and argued that the logics developed by
\citet{McDermott80} and, in the follow-up paper, by
\citet{McDermott82}, are attempts at a logical formalization of 
of autoepistemic statements and not of defaults. Not finding the 
McDermott and Doyle formalisms quite adequate as autoepistemic logics, 
\citet{mo84,mo85} proposed an alternative, the \emph{autoepistemic logic}.

Unfortunately, Moore did not refer to the paper by \citet{re80}
but only to those by \citet{McDermott80} and \citet{McDermott82}, 
and his comments
on this topic were not extrapolated to Reiter's logic.  Neither did Moore 
explain what could go wrong if a default is replaced by
its RMD proposition. Yet, if Moore is right then given the close
correspondence between Reiter's and McDermott and Doyle's views on
defaults, also Reiter's logic is an attempt at a formalization of 
autoepistemic rather than of default propositions. Moreover, if defaults
are really fundamentally different from autoepistemic propositions, as 
Moore claimed, it should be possible to find demonstrable defects of 
Reiter's default logic for reasoning about defaults that could be 
attributed to the different nature of a default and of its Reiter's 
autoepistemic translation. 

Our main objective in Section 
\ref{sec:defaultsvsael} is to 
argue that Moore was right. We show there two forms of such defects
that (1) the RMD proposition is not always {\em sound} in the sense that 
inferences made from it are not always rational with respect to the 
original defaults, and (2) the RMD proposition is not always {\em complete},
that is, there are sometimes rational inferences from the original defaults 
that are not covered by this particular inference rule. In fact, both types 
of problems can be illustrated with examples long known in the literature.  

In the remaining sections, we explain Reiter's default logic as a
formalization of autoepistemic propositions and show that in fact,
Reiter's default logic is a better formalization of Moore's intuitions
than Moore's own autoepistemic logic. On a formal level, our
investigations exploit the results on the unifying semantic framework
for default logic and autoepistemic logic that we proposed earlier
\citep*{DMT03}. That work was based on a algebraic fixpoint theory for
nonmonotone operators \citep*{DeneckerMT00}.  We show that the different
dialects of autoepistemic reasoning stemming from our informal
analysis can be given a principled formalization using these algebraic
techniques. In our overview, we will stress the view on
autoepistemic logic as a logic of {\em autotheoremhood}, in which
theories can include propositions referring to the theory's own
theorems.

\smallskip
\noindent
\textit{Some history.} 
We mentioned that Moore's comments concerning the RMD proposition and
the formalisms by \citet{McDermott80} and \citet{McDermott82}
have never been applied to Reiter's logic. For example,
\citet{Konolige88}, who was the first to investigate the formal link
between autoepistemic reasoning and default logic, wrote that
``\emph{the motivation and formal character of these two systems
  [Reiter's default and Moore's autoepistemic logics -- DMT] are
  different}''. This bypasses the fact that Reiter, as we have seen,
starts his enterprise of building default logic after translating a
default into a proposition which Moore later identified as an
autoepistemic proposition.

There may be several reasons why Moore's comments have never been
extrapolated to Reiter's logic. As mentioned before, one is that Moore
did not refer to the paper by \citet{re80} but only to the papers
by \citet{McDermott80} and \citet{McDermott82}. In addition, the logics 
of Reiter and, respectively,
\MDD were quite different; the formal connection was not known at that
time (mid 1980s) and was established only about five years later
\citep{tru90b}. Also autoepistemic and default logics seemed to be
quite different \citep{mt89a}, and eventually turned out to be
different in a certain precise sense \citep{Gottlob95}. Moreover, the
intuitions underlying the nonmonotonic logics of the time had not been 
so clearly
articulated, not even in Moore's work as we will see later in the
paper, and were not easy to formalize. This was clearly demonstrated
about ten years later by \citet{Halpern97}, who
reexamined the intuitions presented in the original papers of default
logic, autoepistemic logic and Levesque's \citeyearpar{le90}
related logic of only
knowing and showed gaps and ambiguities in these
intuitions, and various non-equivalent ways in which they could be
formalized.

\ignore{
Independently of the above arguments, another reason was that in the
early days of nonmonotonic reasoning, it was thought that the same
formal language could be used to model a number of different knowledge
patterns \cite{McCarthy}: defaults, but also statements such as ``{\em
  Normally A's are B's}'' or ``{\em typically A's are B's}'', frame
axioms, communication agreements, etc. While it was clear to many that
these statements were not equivalent on the informal level, it was
assumed that they would be similar enough to be represented by the
same logic. In this spirit, people cared less about the
non-equivalence of a default and its RMD proposition, even more since
Moore never concretely explained what could go wrong if the first is
replaced by the second.}

\ignore{Finally, Moore distinguished
between defaults and autoepistemic statements, but even he did not
concretely analyze the problems that the use of autoepistemic
propositions to encode defaults brought to Reiter, McDermott and
Doyle's approaches, and this remained unclear, even to this day. }

As a result, the nature of autoepistemic propositions, its
relationship to defaults and what may go wrong when the latter are
encoded by the first, was never well understood.  The relevance of
Moore's claims for Reiter's default logic has never become generally
acknowledged. Reiter's logic has never been thought of and has never
been truly analyzed as a formalization of autoepistemic reasoning. The
influence of Reiter's paper has been so large, that even today, the
default \emph{``most $A$'s are $B$'s''} and the statement \emph{``if
  $x$ is an $A$ and if it is consistent to assume that $x$ is a $B$,
  then $x$ is a $B$''}\footnote{ Or its propositional
  version\emph{``if $A$ and if it is consistent to assume $B$, then
    $B$''}.} are still considered synonymous in some parts of the
nonmonotonic reasoning community. Yet, in fact, they are quite
different and, more importantly, a logical representation of the
second is unsatisfactory for reasoning about the first. 

\section{Reiter's Defaults Are Not Defaults But Autoepistemic Statements}
\label{sec:defaultsvsael}

Our goal below is to justify the claim in the title of the section. To
avoid confusion, we emphasize that by a \emph{default} we mean an
informal expression of the type \emph{most $A$'s are $B$'s}. In
Reiter's approach (similarly in that of \MDD), the default is first
translated into an {\em RMD proposition} \emph{if $x$ is an $A$ and if
  it is consistent with the available information to assume that $x$
  is a $B$, then $x$ is a $B$}, which is then expressed by a
\emph{Reiter's default expression} in default logic:
\[
\default{A(x):M\,B(x)}{B(x)}\; .
\]

To explain the section title, let us assume a setting in which a human
expert has knowledge about a domain that consists of propositions and
defaults.  In the approach of Reiter (the same applies to \MDD), the
expert builds a knowledge base $T$ by including in $T$ formal
representations of the propositions (given as formulas in the language
of classical logic) and of RMD propositions of the defaults (given by
the corresponding Reiter's default expressions). The presence of
Reiter's default expressions in $T$ means that $T$ contains
propositions referring to its own information content, i.e., to what
is consistent with $T$, or dually to what $T$ entails or does not
entail. \citet{mo85} called such reflexive propositions
\emph{autoepistemic} and argued that they statements could be phrased
in terms of theorems and non-theorems of $T$.

Reiter developed a default expression as a formal expression of the
RMD proposition rather than of the default itself (the same holds for
\MDD).  This is why this logic expression does not capture the full
informal content of the default.  When considered more closely, it
indeed becomes apparent that a default and its RMD proposition are not
equivalent or even related in a strict logical sense. A
straightforward possible-world analysis reveals this.  The default
might be true in the actual world (say 95\% of the $A$'s are $B$'s)
but if there is just one $x$ that is an $A$ and not a $B$, and for
which $T$ has no evidence that it is not a $B$, the RMD proposition is
false in this world and $x$ is a witness of this. Thus, it
is obvious that in many applications where a default holds, its RMD
proposition does not.  Conversely, the default might not hold in the
actual world (few $A$'s are in fact $B$'s) yet the expert knows all
$x$'s that are not $B$'s, in which case the RMD proposition is true.

A fundamental difference pointed out by Moore between defaults and
autoepistemic propositions, is that the latter are naturally {\em
  nonmonotonic} but inference rules used for reasoning with them are
not {\em defeasible}. For example, extending the knowledge base $T$
containing an RMD proposition with new information, e.g., that some
$x$ is not a $B$, may indeed have a nonmonotonic effect and delete
some previous inferences, e.g., that $x$ is a $B$. The initial
inference of \emph{$x$ is a $B$}, resulted in a fact that was false.
However, that inference was not defeasible. The essential property of a
defeasible inference is that it may derive a false conclusion from 
premises that are true in the actual world. For instance, the inference 
from \emph{most $A$'s are $B$'s} and \emph{$x$ is an $A$}
that \emph{$x$ is a $B$} is defeasible as its consequent may be false
while the premises are true. In the context of our example above
the theory, say $T$, entailed the false fact that \emph{$x$ is a $B$} 
from the premises (i) the RMD proposition, (ii) $x$ is an $A$ and (iii)
$T$ contained no evidence that $x$ is not a $B$. It was not defeasible
since one of its premises was false. Indeed, the RMD proposition was 
false and $x$ was a witness. The inference rules applied are not 
defeasible (they are, essentially, the introduction of conjunction and 
modus ponens). To sum up, an inference from a knowledge base involving 
an RMD proposition may be false but only if the RMD proposition itself 
is false.

To emphasize further consequences of equating defaults and RMD propositions 
we will look at well-known examples from the literature. First, we turn our
attention to the question whether there are cases when applying the RMD 
proposition leads to inferences that do not seem rational (lack of 
``soundness'' with respect to understood informally ``rationality''). The 
Nixon Diamond example by \citet{ReiterCriscuolo81} and reasoning problems with 
related inheritance networks illustrate the problems that arise.

\begin{example}
{\rm
  Richard M. Nixon, the 37th president of the United States, was a
  Republican and a Quaker. Most Republicans are hawks while most
  Quakers are doves (pacifists). Nobody is a hawk and a dove. Some people are
  neither hawks nor doves. Encoding the Reiter-McDermott-Doyle
  proposition of these defaults in default logic, we obtain the
  following theory:
\[ \begin{array}{c} Republican(Nixon) \land Quaker(Nixon)\\
  \medskip
  \forall x (\neg Dove(x)\lor \neg Quaker(x))\\
  \default{Republican(x) :M\ Hawk(x)}{ Hawk(x)} \hspace{1cm}
  \default{Quaker(x) :M\ Dove(x)}{Dove(x)}\; .
\end{array}
\] 
In default logic, this theory gives rise to two extensions. In one of
them Nixon is believed to be a hawk and not a dove, in the other one,
a dove and not a hawk. But is this rational? As we mentioned above,
the use of an RMD-proposition is rational when it is expected to hold
for most $x$, and hence, in absence of information, it is likely to
hold for some specific $x$. But in the case of Nixon, we know in
advance that at least one of the two ``Nixon'' instances of the RMD
propositions has to be wrong. As to which one is wrong, without
further information one could as well throw a coin. Moreover, it is
not unlikely that they are both wrong and that in fact, Nixon is
neither dove nor hawk. And in fact, it seems more rational not to
apply any of the defaults, leading to a situation where it is not
known whether Nixon is a dove, a hawk or neither. The rationale of
using the RMD proposition as a substitute for the
default 
does not hold for Nixon or any other
republican quaker for that matter. 
\hfill$\Box$
}
\end{example}

\begin{example} 
{\rm
  Let us assume now that all quakers are republicans. In this case, the 
  default that most quakers (say 95\%)
  are doves is more specific than and overrules the default that most
  republicans (say 95\%) are hawks. It is rational here to give
  priority to the quaker default, leading to the defeasible conclusion
  that Nixon is a dove. However, this conclusion cannot be derived from 
  the RMD propositions because their consistency premise \emph{``it is
  consistent to assume that $x$ is a dove (respectively a hawk)''}
  is too general to take such information into account.
\hfill$\Box$
}
\end{example}

Such scenarios were studied in the context of inheritance hierarchies
\citep{tzky86}.  To reason correctly on this sort of applications using
Reiter's logic, the consistency condition of the RMD propositions has
to be tweaked to take the hierarchy into account and give priority to
the quaker default. For example, we can reformulate the RMD
proposition of the default \emph{``most republicans are hawks''} as
\emph{``if x is known to be a republican and it is consistent to
  assume that he is a hawk and it is consistent to assume that he is
  not a quaker, then $x$ is a hawk''}, which takes additional
information into account.  Such modified rules can of course be
represented in default logic. After all, the logic was developed for
representing (defeasible) inference rules.  But, as in the examples
above, they cannot be \emph{inferred} from the RMD propositions. And
the inferences that can be drawn from the RMD propositions are not
always the rational ones.

The next problem that arises is of a complementary nature and concerns
(lack of) completeness with respect to ``rational'' inferences. Are
there cases where rational albeit defeasible inferences can be drawn
from defaults that cannot be inferred from RMD propositions? As
suggested above by our general discussion, the answer is indeed
positive. After all, the RMD proposition expresses only a single and
quite specific type of inference that might be associated with a
default.

\ignore{Given that the RMD proposition expresses {\em one} defeasible
inference rule related to its default,  question (2) is whether this
single rule {\em exhausts} the defeasible inferences that can be made
from the original default. That is, are there other (defeasible) inferences
that could be drawn from the default but not from its RMD proposition?}

\begin{example}
{\rm
As an illustration, let us consider the defaults \emph{most Swedes are
  blond} and \emph{most Japanese have black hair}. Nobody is both Swede and
Japanese, or has both blond and black hair. If we learn
know that Boris is a Swede or a Japanese then, given that he cannot be
both Swede and Japanese, it seems rational  to conclude defeasibly
that Boris's hair is blond or black. In other words, defaults
can (sometimes) be combined and together give rise to defeasible
inference rules like: 

\[
\default{\mbox{\it{Boris is Swede or Japanese: M Boris's hair is blond or 
black}}}{\mbox{{\it Boris's hair is blond or black}}}\; {.}
\]

If all we know is that Boris is Swede or Japanese, the conclusion
of this rule cannot be drawn from the two original RMD propositions
for the simple reason that for each, one of their premises is not
satisfied: it is not known that Boris is a Swede, and neither is it
known that he is Japanese.  For instance, in 
the logic of Reiter, the two defaults would be encoded as
\[
\default{Swede(x): M\,Blond(x)}{Blond(x)}\quad\mbox{and}\quad
\default{Japanese(x): M\, Black(x)}{Black(x)}\; \mbox{.}
\]
If we only know $Swede(Boris) \lor Japanese(Boris)$, then neither 
$Swede(Boris)$ nor $Japanese(Boris)$ can be established. 
Therefore, the premises of neither rule are established and no inference can
be made.  Even more, if we accept Reiter's logic as a logic of
autoepistemic propositions, these conclusions {\em should not be
  drawn} from these expressions.
}
\hfill$\Box$
\end{example}

This example shows a clear case of a desired defeasible inference that
cannot be drawn from the rules expressed in the two RMD
propositions. A default expression in Reiter's logic that would do the job 
has to encode explicitly the combined inference rule:
\[
\default{Swede(x)\lor Japanese(x): M(Blond(x)\lor Black(x))}{Blond(x)
\lor Black(x)}\; .
\]
This expresses an inference rule which is not derivable from
the original RMD propositions in the logics of Reiter, McDermott,
Doyle, or Moore.  Default logic does not support such reasoning unless
the combined inference rule is explicitly encoded as well.

\begin{example}
{\rm
Assume that we now
find out that Boris has black hair. Given that he is Japanese or
Swede, and given the defaults for both, it seems rational to assume
that he is Japanese. Can we infer this from the combined inference
rules expressed above and given that nobody can be blond and black, or
Swede and Japanese? The answer is no and, consequently, yet another
inference rule should be added to obtain this inference.
}
\mbox{\ }\hfill$\Box$
\end{example}

\ignore{
In the context of natural language, another famous inference made from
defaults are those given by {\em Gricean implicatures}.  When we say
``\emph{Most Swedes are blond}'' we are often implying that some
Swedes are not. This inference is a \emph{scalar implicature}
generated by the \emph{Gricean Maxim of Quantity} \cite{grice89}. 
It is rooted in the conversational convention that expressing a
statement in some scalar order (in our case: no Swede is blond, some
Swede is Blond, most Swedes are blond, all Swedes are blond), normally
entails the falsity of later elements in this order.  Neither in
Reiter's default logic nor in the logics of McDermott and Doyle the
RMD proposition for the default \emph{``most Swedes are blond''}
entails the statement \emph{``there is a Swede that is not
  blond''}.\footnote{As a remark, we probably want to apply the scalar
implicature derived from ``\emph{Most Swedes are blond}'' only when
reasoning about large groups of Swedes.  It may be unsafe to apply
this inference rule in a setting involving only small groups of
Swedes, as then there might well be no exceptions to the default.}
}

Problems of these kind were reported many times in the NMR literature 
and prompted attempts to ``improve'' Reiter's default logic so as  to
capture additional defeasible inferences of the informal default. 
This is, however, a difficult enterprise, as it starts from a logic
whose semantical apparatus is developed for a very specific form of
reasoning, namely autoepistemic reasoning. And while at the formal level
the resulting logics \citep{br91,schau92a,luka88,mitru94} capture some 
aspects of defaults that Reiter's logic does not, also they formalize a 
small fragment only of what a default represents and, certainly, none 
has evolved into a method of reasoning about defaults. In the same time, 
theories in these logics entail formulas that cannot be justified from the
point of view of default logic as an autoepistemic logic.

\ignore{Finally, we note that all differences and misalignments aside, in some 
cases RMD propositions may be acceptable substitutes for defaults but 
only when there is a reasonable expectation that for \emph{most} $x$'s 
the RMD proposition holds: if $x$ is known to be an $A$ and not known to 
be a non-$B$, then it is a $B$. If there are no reasons to believe that 
most instances of the RMD proposition are true, it would be irrational 
to use this axiom.}

To summarize, an RMD proposition expresses one defeasible inference
rule associated with a default. It often derives rational assumptions
from the default but not always, and it may easily miss some useful
and natural defeasible inferences. The RMD proposition is
autoepistemic in nature; Reiter's original default logic is therefore a
formalism for autoepistemic reasoning. As a logic in which inference
rules can be expressed, default logic is quite useful for reasoning on
defaults. The price to be paid is that the human expert is responsible 
for expressing the desired defeasible inference rules stemming from the 
defaults and for fine-tuning the consistency conditions of the inference
rules in case of conflicting defaults. This may require substantial 
effort and leads to a methodology that is not elaboration tolerant.

While our discussion shows that in general, RMD propositions and
Reiter's defaults do not align well with the informal concept of a 
default of the form \emph{most $A$'s are $B$'s}, there are other 
nonmonotonic reasoning patterns that are correctly expressed through 
Reiter's defaults. In particular, patterns such as communication 
conventions, database or information storage conventions and policy rules
in the typology of \citet{McCarthy86}, can be expressed well by {\em true}
autoepistemic propositions and, consequently, are correctly formalized
in Reiter's logic. E.g., the convention that an airport customs database 
explicitly contains the nationality of only non-American passengers, is
correctly specified by  the Reiter default $$\default{:M
Nationality(x)=USA}{Nationality(x)=USA}\; \hbox{.}$$ Similarly, the policy rule 
that the departmental meetings are normally held on Wednesdays at noon,
is correctly formalized by
$$\default{:M Time(meeting)="Wed,\ noon"}{Time(meeting)="Wed,\ noon"}\;
\hbox{.}$$ 
%

\ignore{Importantly,
they do not illustrate problems of default logic seen as an
autoepistemic logic; rather they are consequences of the mismatch
between a default and its RMD proposition.
In the following sections of the paper, we show that Reiter's default
logic is not only a logic of autoepistemic reasoning but a
particularly important one.}

\ignore{
There are many more such examples. But we feel our points are clear: 
defaults are not equivalent with their RMD propositions, RMD propositions are  autoepistemic in nature, Reiter's formal ``defaults''
do not represent defaults, and Reiter and McDermott-Doyle logics 
capture autoepistemic propositions. These logics provide a way to 
express some {\em defeasible inference rules} associated to defaults.
In the following sections of the paper, we show that Reiter's default
logic is not only a logic of autoepistemic reasoning but a particularly
important one.}

In spite of such examples, the fact remains that default logic is not a logic of 
defaults. 
Are there other logics that could be regarded as such? There have been 
several interesting attempts at formalizing defaults 
\emph{most $A$'s a re $B$'s}. Most important of them focused on defaults
as \emph{conditional assertions} and on abstract
nonmonotonic consequence relations \citep*{mak88,leh89,pea90,klm90,lm92}.
This research direction resulted in elegant mathematical theories and
deep insights into the nature of some forms of nonmonotonic reasoning.
However, it is not directly related to our effort here. Thus, rather
than to discuss it we refer to the papers we cited. 

\ignore{
This prompted researchers to search for alternative formal accounts.
One direction was to try to ``improve'' Reiter's default logic to
capture more of defeasible inferences of the informal default. This is
a difficult enterprise, as it starts from a logic whose semantical
apparatus is developed for autoepistemic reasoning. And while at the
formal level the resulting logics \cite{br91,schau92a,luka88,mitru94}
capture some aspects of defaults that Reiter's logic does not, also
they formalize a small fragment only of what a default represents and,
certainly, none has evolved into a method of reasoning about defaults.
In the same time, theories in these logics entail formulas that cannot
be justified from the point of view of autoepistemic reasoning.
}

Instead, in the remainder of the paper, we focus on the second objective
identified in the introduction. That is, we provide an informal basis to 
autoepistemic reasoning, we place Reiter's
default logic firmly among dialects of autoepistemic reasoning, and show 
that Reiter's logic was a watershed point that pinpointed one of the most 
fundamental and most important forms of autoepistemic reasoning.

\section{Studies of Relationships Between Default Logic and Autoepistemic Logic}
\label{sec:moredefaultsvsael}

\ignore{
In his paper, Reiter quite explicitly \emph{identified} a default
\emph{``most A's are B's}'' with its RMD proposition, saying on p. 82:
\begin{quote}
`We take it [that is, the default ``Most birds can fly''
 ---DMT 
] to mean something like ``If an $x$ is a bird,
  then in the absence of any information to the contrary, infer that $x$
  can fly.''' 
\end{quote}
While this statement makes it rather clear that Reiter
was interested only in modeling one particular inference pattern
associated with a default, it seems that it is precisely this
statement that contributed to later confusion about the nature of the
two informal expressions.
}

\citet{Konolige88} was the first to investigate a formal link between default and autoepistemic logic. He
proposed the following translation ${\Kon}$ from default logic to 
autoepistemic logic: 
\[
\default{\alpha : M \beta_1, \dots, \beta_n}{\gamma}\qquad\mapsto\qquad K\alpha
\land \neg K\neg\beta_1\land \dots \land \neg K\neg\beta_n \rightarrow \gamma
\]
and argued that ${\Kon}$ was equivalence preserving in the sense that
default extensions of the default theory were exactly the
autoepistemic expansions of its translation. This translation is
intuitively appealing, essentially expressing formally the RMD
proposition of the default in modal logic, and it indeed plays an
important role in the story. Nevertheless, it turned out that this
translation was only partially correct \citep{koerratum}.
Later, \citet{Gottlob95} presented a correct 
translation from default logic to autoepistemic logic but also proved
that no \emph{modular} translation exists. The latter result showed
that these two logics are essentially different in some important 
aspect. As a result, the autoepistemic nature of default logic, which 
Moore had implicitly pointed at, and his implicit criticism on default 
logic as a logic of defaults were never widely acknowledged. 


But Reiter's logic is just that --- a logic of autoepistemic
reasoning.  Moreover, in many respects it is a better logic of
autoepistemic reasoning than the one by Moore. Our goal now is to
reconsider the intuitions of autoepistemic reasoning, to distinguish
between different dialects of it and to develop principled
formalizations for these dialects. In particular, we relate Reiter's
and Moore's logics, and explain in what sense 
Reiter's
logic is better than Moore's.  Our discussion uses the formal results
we developed in an earlier paper \citep{DMT03}. There we used the 
algebraic fixpoint theory for arbitrary lattice operators 
\citep{DeneckerMT00} to define four different semantics of default 
logic and of autoepistemic logic. This theory can be summarized as 
follows.

A complete lattice $\langle L,\leq\rangle$ induces a complete bilattice 
$\langle L^2,\leqp\rangle$, where $\leqp$ is the precision order
on $L^2$ defined as follows: $(x,y)\leqp (u,v)$ if $x\leq u$ and $v\leq
y$. Tuples $(x,x)$ are called exact. For any $\leqp$-monotone operator $A:L^2\ra L^2$ that is {\em
  symmetric}, that is, $A(x,y)=(u,v)$ if and only if $A(y,x)=(v,u)$, we can define
three derived operators. These four operators identify four different
types of fixpoints or least fixpoints (when the derived operator is
monotone). They are summarized in Table \ref{tab:DMT1} (where the
operator $A_1(\cdot,\cdot)$ used to define $O_A$ is the projection of
$A$ on the first coordinate).

\begin{table}
\cent{\begin{tabular}{lll}
    \wis{A:L^2\ra L^2} &  & \quad{Kripke-Kleene least fixpoint}\\
    \wis{O_A:L\ra L} & \wis{O_A(x) =  A_1(x,x)} & \quad{Supported fixpoints}\\
    \wis{S_A:L\ra L} & \wis{S_A(x) = \mathit{lfp}(A_1(\cdot,x))} & \quad{Stable fixpoints}\\
    \wis{\mathcal{S}_A:L^2\ra L^2} &
    \wis{\mathcal{S}_A(x,y)=(S_A(y),S_A(x))} & \quad{Well-founded least
      fixpoint}
  \end{tabular}}
\caption{Lattice operators and the corresponding semantics}
\label{tab:DMT1}
\end{table}

By assumption, $A$ is a $\leqp$-monotone operator on $L^2$ and its
$\leqp$-least fixpoint is called the \emph{Kripke-Kleene} fixpoint of $A$.
Fixpoints of the operator $O_A$ correspond to exact fixpoints of $A$
($x$ is a fixpoint of $O_A$ if and only if $(x,x)$ is a fixpoint of $A$)
and are called \emph{supported} fixpoints of $A$. The operator \wis{S_A}
is an anti-monotone operator on $L$. Its fixpoints yield exact fixpoints
of $A$ (if $x$ is a fixpoint of $S_A$ then $(x,x)$ is a fixpoint of $A$).
They are called \emph{stable} fixpoints of the operator $A$. It is clear
that stable fixpoints are supported.
The operator
\wis{\mathcal{S}_A} is a $\leqp$-monotone operator on $L^2$ and its
$\leqp$-least fixpoint is called the well-founded fixpoint of $A$ (fixpoints
of \wis{\mathcal{S}_A} are also fixpoints of $A$). The names of these
fixpoints reflect the well-known semantics of logic programming, where
they were first studied by means of operators on lattices. 
Taking Fitting's four-valued immediate
consequence operator \citep{Fitting85} for $A$, we proved \citep{DeneckerMT00}
that the four different types of fixpoint correspond
to four well-known  semantics of logic programming: Kripke-Kleene
semantics \citep{Fitting85}, supported model semantics \citep{Clark78},
stable semantics \cite{Gelfond88} and well-founded semantics
\citep{Vangelder91}.

This elegant picture extends to default logic and autoepistemic
logic \cite{DMT03}. In that paper, we identified the semantic operator
\wis{{\mathcal E}_\Delta} for a default theory $\Delta$, and the 
semantic operator \wis{\D_T} for an autoepistemic theory $T$. Both 
operators where defined on the bilattice of possible-world sets, which we
introduce formally in the following section. Just as for logic programming, 
each operator determines three derived operators and so, for each logic
we obtain four types of fixpoints, each inducing a semantics. Some
of these semantics turned out to correspond to semantics proposed earlier;
other semantics were new. Importantly, it turned out that the operators
\wis{{\mathcal E}_\Delta} and \wis{\D_{\Kon(\Delta)}} are identical.
Hence, Konolige's mapping turned out to be equivalence preserving for {\em
  each} of the four types of semantics! Table \ref{tab:DMT2}
summarizes the results. The first two lines align the theories and the
corresponding operators. The last four lines describe the matching
semantics (the new semantics for autoepistemic and default logics
obtained from this operator-based approach \cite{DeneckerMT00}
are in bold font).

\begin{table}
\cent{ \begin{tabular}{lll}
default theory \wis{\Delta} & \onKon   &  \quad autoepistemic theory \wis{T} \\
semantic operator \wis{{\mathcal E}_\Delta} & \onKon & \quad semantic operator \wis{\D_T}\\
{\bf KK-extension}  & \onKon  & \quad {KK-extension}\\
 & & \quad {\footnotesize \citep{DMT98}}\\
{Weak extensions} & \onKon  & \quad {{Moore expansions}}\\
{\footnotesize \citep{mt89a}} && \quad {\footnotesize \citep{mo84}} \\
    {{Reiter extensions}} & \onKon  &  \quad {{\bf Stable extensions}}\\
    {\footnotesize \citep{re80}} & & \\
    {Well-founded extension}\mbox{\quad} & \onKon &
    \quad {{\bf Well-founded extension}}\\
    {\footnotesize \citep{Baral91}} & & 
  \end{tabular}}
\caption{The alignment of default and autoepistemic logics}
\label{tab:DMT2}
\end{table}

From this purely mathematical point of view Konolige's intuition seems
basically right. His mapping failed to establish a correspondence
between Reiter extensions and Moore expansions \emph{only} because they are on
different levels in the hierarchy of the semantics. Once we correctly align the
dialects, his transformation works perfectly.  Conversely, we also
proved that the standard method to eliminate nested modalities in the
modal logic S5 can be used to translate any autoepistemic logic theory
$T$ into a default theory that is equivalent to $T$ under each of the
four semantics.

While the non-modularity result by \citet{Gottlob95} had shown that
default logic and autoepistemic logic are essentially different
logics, our results summarized above unmistakenly point out that
default and autoepistemic logics are tightly connected logical
systems. They suggest that the four semantics formalize
different {\em dialects} of autoepistemic reasoning and that Reiter
and Moore formalized different dialects. Therefore, in the rest of the
paper, we will view Reiter's logic simply as a fragment of modal logic,
as identified by Konolige's mapping. 

\section{Formalizing Autoepistemic Reasoning --- an Informal Perspective}
\label{informal}

In our paper \citep{DMT03} we developed a purely algebraic, abstract study of
semantics. The study identified the (nonmonotone) operators of autoepistemic
and default logic theories, and applied the different notions of
fixpoints to them. What that paper was missing was an account of what
these fixpoint constructions mean at the informal level and how the
different dialects in the framework differ. Being as clear as possible 
about the informal semantics of autoepistemic theories is essential, 
as it is there where problems with formal accounts start. 

This is the gap that we close in the rest of this paper. To this end 
we first return to the original concern of Reiter, and of McDermott and 
Doyle. Let us suppose that we have incomplete knowledge about the actual
world, represented in, say, a first order theory $T$, and that we know
that most $A$'s are $B$'s. Following the Reiter, \MDD approach, we 
would like to assert the following proposition: 
\begin{quote}
If for some $x$, $T\models
  A(x)$ and $B(x)$ is consistent with $T$ (that is, $T\not\models \neg
  B(x)$), then $B(x)$. 
\end{quote}
In fact, we would like to express this statement
{\em in} the logic and, moreover, to add this proposition, with its
references to what $T$ entails or does not entail, to $T$ itself. What
we obtain is a theory $T$ that refers to its own theorems. In this
view then, modal literals $K\varphi$ in an autoepistemic theory $T =
\verz{ \dots F[K\varphi]\dots } $ 
are to be
interpreted \emph{informally} as statements $T\models \varphi$, and
the theory $T$ itself as having the form $T = \verz{ \dots
  F[T\models \varphi] \dots}$, emphasizing the intuition of the
self-referential nature of autoepistemic theories.

This view reflects what seems to us the most precise intuition that Moore 
proposed: to view autoepistemic propositions as inference rules. Specializing
the discussion above to the autoepistemic formula
\begin{equation}
\label{eq:AELprop} 
K \alpha_1 \land \dots \land K \alpha_n \land \neg K \beta_1 \land \dots 
\land \neg K \beta_m \rightarrow \gamma
\end{equation}
we can write it (informally) as:
\[ 
T \models \alpha_1 \land \dots \land T\models \alpha_n \land T \not 
\models \beta_1 \land \dots \land  T \not \models  \beta_m \rightarrow 
\gamma,
\]
and understand it (informally) as an inference rule:
\begin{eqnarray}
\label{eq:AELinfrule}
  && \mbox{if $\alpha_1, \ldots, \alpha_n$ are theorems and
  $\beta_1,\ldots,\beta_m$ are {\em not} theorems}\\
  && \mbox{then $\gamma$ holds.\ignore{is a theorem.}} \nonumber
\end{eqnarray}
which is consistent with Moore's \citeyearpar[p. 76]{mo85} position
we cited earlier.
\ignore{In Moore's view, a modal literal $K\varphi$ was interpreted informally
as ``\emph{$\varphi$ is a theorem}''. But a theorem of what? The
answer was: a theorem of the autoepistemic theory $T$, of which the
proposition is a part.  Hence, a rule such as (\ref{eq:AELprop}) is
understood \emph{informally} as:
\[ 
T \models \alpha_1 \land \dots \land T\models \alpha_n \land T \not 
\models \beta_1 \land \dots \land  T \not \models  \beta_m \rightarrow 
\gamma,
\]
} 
Alternatively, $K\varphi$ can be read as ``\emph{$\varphi$ can be
  derived, or proven}'' (again, from the theory itself), which amounts 
at the informal level just to a different wording. We will refer to this
notion of theorem and derivation as \emph{autotheorem} and
\emph{autoderivation}, respectively. Accordingly, we will call the
basic Moore's perspective as that of {\em autotheoremhood}. 

The autotheoremhood view can be seen as a special case of a more
generic view, also proposed by Moore, based on \emph{autoepistemic
  agents}. In this view which, incidentally, is the reason behind the
name {\em autoepistemic logic}, an autoepistemic theory is seen as a
set of introspective propositions, believed by the agent, about the
actual world and his own beliefs about it.  The crucial assumption is
the one which \citet{le90} dubbed later the \AIKA: the
assumption that all that is known by the agent is {\em grounded} in
his theory, in the sense that it belongs to it or can be derived from
it. In the case of the autotheoremhood view, the agent is nothing else
than a personification of the theory itself, and what it knows is
what it entails.  We discuss alternative instances of this agent-based
view in the next section.

But let us now focus on developing the autotheoremhood perspective. We
regard it as a more precise intuition that is more amenable to formalization 
despite the fact that self-reference, which is evidently
present in the notion of autotheoremhood, is a notoriously complex 
phenomenon. It plagued, albeit in a different form,  the {\em theory of 
truth} in philosophical logic with millennia-old paradoxes 
\citep{Tarski33,Kripke75,BarwiseEtchemendy87}. The
best known example is the famous {\em liar} paradox: 
\cent{{\em ``This sentence is false.''}} 
An autoepistemic theory that is clearly reminiscent of this paradox is: 
\[ 
T_{\mathit{liar}} = \verz{ \neg K P \mim P}. 
\] 
In the autotheoremhood view, this theory states that if \emph{it} 
does not entail $P$ then $P$ holds. However, if $P$ is not entailed, 
then we \emph{have} an argument for $P$, and if $P$ is entailed, the 
unique proposition of the theory is trivially satisfied; no argument 
for $P$ can be constructed. This is \emph{mutatis mutandis} the argument
for the inconsistency of the liar sentence. In view of the difficulties 
that self-reference has posed to the development of the theory of truth, 
it would be naive to hope that a crisp, unequivocal formalization of 
autoepistemic logic existed. 

\citet[p. 82]{mo85} explained the difficulty of defining the semantics for 
autoepistemic inference rules (\ref{eq:AELinfrule}) as follows. 
When the inference rules are monotonic, that is,
when $m=0$, 
\begin{quote}
`once a formula has been generated at a given stage, it 
remains in the generated set of formulas at every subsequent stage. 
[...] The problem with attempting to follow this pattern with 
nonmonotonic inference rules [that is, when $m>0$ (note of the authors)]
is that we cannot draw nonmonotonic inferences reliably at any 
particular stage, since something inferred at a later stage may
invalidate them.'
\end{quote}
To put it differently, the problem is that when a 
rule (\ref{eq:AELinfrule}) is applied to derive $\gamma$ at some stage when all $\alpha_i$'s 
have been inferred to be theorems and none of the $\beta_j$'s has been 
derived, later inferences may derive some $\beta_j$ and hence, 
invalidate the derivation of $\gamma$. In such case, Moore argues, all 
we can do is to characterize the desired result as the solution of a 
\emph{fixpoint equation} instead of computing it by a \emph{fixpoint 
construction}:
\begin{quote}
 `Lacking such an iterative structure, nonmonotonic systems often
  use nonconstructive ``fixed point'' definitions, which do not directly
  yield algorithms for enumerating the ``derivable'' formulas, but do
  define sets of formulas that respect the intent of the nonmonotonic
  inference rules.' 
\end{quote}
This was an extremely clear and compelling 
representation of intuitions behind not only the Moore's own autoepistemic
logic, but also behind the formalisms of McDermott and Doyle, and of Reiter, 
too, for that matter.

It is useful now to look at these ideas from a more formal point of 
view. Let us consider a modal theory $T$ over some vocabulary $\Sigma$.
Let $T$ consist of ``inference rules'' of the form (\ref{eq:AELinfrule}), 
where for simplicity we assume that all formulas $\alpha_i,
\beta_j, \gamma$ are objective (that is, contain no modal 
operator).\footnote{Our approach works equally well for
arbitrary modal theories.} 
The inference processes that Moore had in mind are syntactic in nature 
and are derivations of formulas. Yet, it is straightforward to cast 
these inference processes in semantical terms. 

Let $\Worlds$ be the set of all $\Sigma$-interpretations. A state of 
belief is represented as a set  $\bs \subseteq \Worlds$ of possible 
worlds.\footnote{A \emph{possible-world set} is a special Kripke
structure in which the accessibility relation is total.}
Intuitively, each element $w\in\bs$ represents a {\em possible
world}, a state of affairs that satisfies the agent's beliefs. A world 
$w\not\in\bs$ represents an {\em impossible world}, a state of affairs 
that violates at least one proposition of the agent. 
Given a set $\bs$ representing the worlds held possible by an agent,
the following, standard, definition formalizes which (modal) formulas
the agent believes (or knows --- we do not distinguish between these two
modalities in our discussion).
\begin{definition} We define the satisfiability relation $\bs,w\models
  \varphi$ as in the modal logic S5 by the standard recursive
  rules of propositional satisfaction augmented with one additional rule:
\[
\bs,w \models K\varphi \mbox{ if for every } v\in\bs, \bs,v\models 
\varphi. 
\]
We then define $\bs \models K\varphi$ ($\varphi$ is believed or known in state
$B$) if for every $w\in \bs$, $\bs,w\models \varphi$.
\end{definition}
This definition extends the standard definition of truth in the sense
that if $\varphi$ is an objective formula then $\bs,w\models\varphi$
if and only if $w\models\varphi$. We define $\Theo(\bs) = \{\varphi
\st \bs \models K\varphi\}$ and $\Theo_{obj}(\bs)$ the restriction of
$\Theo(\bs)$ to objective formulas. These sets represent 
all modal formulas and all objective formulas, respectively, known in 
the state of belief $\bs$.  

It is natural to order belief states according to ``how
much'' they believe or know. For two belief states $\bs_1$ and
$\bs_2$, we define $\bs_1\leq_k\bs_2$ if $\Theo_{obj}(\bs_1) \subseteq
\Theo_{obj}(\bs_2)$ or, equivalently, if $\bs_2\subseteq \bs_1$. The 
ordering $\leq_k$ is often called the \emph{knowledge} ordering. We
observe that $\bs_1\leq_k\bs_2$ does not entail $\Theo(\bs_1)
\subseteq \Theo(\bs_2)$, due to the nonmonotonicity of modal literals
$\neg K \varphi$ expressing ignorance, some of which may be true in $\bs_1$
and false in $\bs_2$.

We can see Moore's
inference processes as sequences $(\bs_i)_{i=0}^\lambda$ of
possible-world sets such that $\bs_0=\Worlds$, the possible-world set
of maximum ignorance in which only tautologies are known. In each
derivation step $\bs_i\ra\bs_{i+1}$, some worlds $w\in\bs_i$ might be
found to be impossible and eliminated in $\bs_{i+1}$; other worlds
$w\not\in\bs_i$ might be established to be possible and added to
$\bs_{i+1}$. This process is described through Moore's semantic
operator $D_T$, which maps a possible-world set $\bs$ to the
possible-world set $\{ w \st \bs,w\models T\}$. For theories
consisting of formulas (\ref{eq:AELprop}), $D_T(\bs_i)$ is exactly the
set of all possible worlds that satisfy the conclusions $\gamma$ of
all inference rules that are ``active'' in $\bs_i$, that is, for which
$\bs_i\models K\alpha_j$, $1\leq j\leq n$, and $\bs_i\not \models
K\beta_j$, $1\leq j\leq m$.

Let us come back to Moore's claims. The nonmonotonicity of the
inference rules (\ref{eq:AELinfrule}), or more precisely, formulas
(\ref{eq:AELprop}) is due to the negative conditions 
$\neg K\beta_j$ ($\beta_i$ not known, not proved, not a theorem). So let us
assume that $m=0$ for all inference rules in $T$.\footnote{For arbitrary 
theories $T$, the corresponding assumption is that there are no modal 
literals $K\varphi$
occurring positively in $T$.} One can show that under this assumption
$D_T$ is a monotone operator with respect to $\subseteq$: if $\bs_1
\subseteq\bs_2$, then $D_T(\bs_1)\subseteq D_T(\bs_2)$. This can be 
rephrased in terms of knowledge ordering: if $\bs_1\leq_k\bs_2$, then
$D_T(\bs_1)\leq_k D_T(\bs_2)$. In other words, the operator $D_T$ is
also monotone in terms of the knowledge ordering $\leq_k$. Moore's
inference process $(\bs_i)_{i=0}^\lambda$ is now an \emph{increasing}
sequence in the knowledge order $\leq_k$. It yields a least fixpoint 
$\bs_T$ in the knowledge order (equivalently, the greatest fixpoint of 
$D_T$ in the subset order $\subseteq$). Every other fixpoint of $D_T$ 
contains more knowledge than $\bs_T$. The fixpoint $\bs_T$ is the
intended belief state associated with the theory $T$ of monotonic 
inference rules.

In the general case of nonmonotonic inference rules
($m>0$, for some rules), the operator $D_T$ may not be
monotone. The inference process constructed with $D_T$ may oscillate
and never reach a fixpoint, or may reach an unintended fixpoint due to
the fact that it may derive that a world is impossible on the basis of
an assumption $\neg K\beta_i$ which is later withdrawn. In such
case, stated Moore, all we can do is to focus on possible-world sets
that ``respect the intent of the nonmonotonic inference rules'' as
expressed by a {\em fixpoint equation} associated to $T$, rather than
being the result of a {\em fixpoint construction}. In this way Moore
arrived at his semantics of autoepistemic logic, summarized in the
following definition.

\begin{definition} An autoepistemic expansion of a modal theory $T$
  over $\Sigma$ is a possible-world set $\bs\subseteq \Worlds$ such
  that $\bs=D_T(\bs)$. 
\end{definition}

We agree with Moore that the condition of being a fixpoint of $D_T$ is
a necessary condition for a belief state to be a possible-world model 
of $T$. However, it is obviously not a sufficient one, at least not in 
the autotheoremhood view on $T$. This is  obvious, as this 
semantics does not coincide with Moore's own ideas on the semantics 
of monotonic inference rules. A counterexample is the following theory:
\[
T = \verz{ K P \mim P}.
\] 
This theory consists of a unique monotonic inference rule, albeit a
rather useless one as it says ``\emph{if $P$ is a theorem then $P$
  holds}''. According to Moore's account of monotonic inference rules,
the intended possible-world model of this theory is
$\Worlds=\verz{\emptyset, \{P\}}$ (we assume that $\Sigma=\{P\}$). Yet,
$T$ has two autoepistemic expansions, the second being the
self-supported possible-world set $\verz{\{P\}}$.

It is worth noting that this theory is related to yet another
famous problematic statement in the theory of truth, namely the {\em
  truth sayer}: \cent{\em ``This sentence is true.''} The truth value
of this statement can be consistently assumed to be true, or equally
well, to be false. Therefore, in Kripke's \citeyearpar{Kripke75} 
three-valued truth theory, the truth
value of the truth sayer is {\em undetermined} $\Un$. In case of the
related autoepistemic theory $\{KP \rightarrow P\}$, also Moore's
semantics does not determine whether $P$ is known or not. But in the
autotheoremhood view, it is clear that $P$ should not be known and
this transpires from Moore's own explanations on monotonic inference
rules.\footnote{There does not seem to be an analogous strong argument
  why the truth sayer sentence should be false. Yet,
  \citet{Fitting97} proposed a refinement of Kripke's theory of
  truth in which truth is minimized and the truth sayer statement is
  {\em false}. For this, he used the same well-founded fixpoint
  construction that we will use below to obtain a semantics that
  minimizes knowledge for autotheoremhood theories.} We come back to
the issue of self-supported expansions in Section~\ref{sec:other},
where we explore alternative perspectives on autoepistemic
propositions, in which such self-supported expansions might be
acceptable.

The main question then is: Can we improve Moore's method to build
inference processes in the presence of nonmonotonic inference rules in
$T$? In this respect, the situation has changed since 1984. The
algebraic fixpoint theory for nonmonotone lattice operators 
\citep{DeneckerMT00},
which we developed and then used to build the unifying semantic
framework for default and autoepistemic logics \citep{DMT03}, gives us 
new tools for defining fixpoint 
constructions and fixpoint equations which can be applied to Moore's
problem. 

We illustrate now these tools in an informal way and refer to 
these intuitions later when we introduce major concepts for a formal 
treatment. Let us consider the theory:
\[
T = \verz{ P,\; \neg K P \mim Q ,\; K Q \mim Q }. 
\]
Informally, the
theory expresses that $P$ holds, that if $P$ is not a theorem then $Q$
holds, and that if $Q$ is a theorem, then $Q$ holds. Intuitively, it
is  clear what the model of this theory should be: $P$ is a
theorem, hence the second formula cannot be used to derive $Q$ and
neither can the truth sayer proposition $K Q\mim Q$. Therefore, the
intended possible-world set is $\bs_T=\{\{P\},\{P,Q\}\}$, that is, $P$ is
entailed, $Q$ is unknown.

It is easily verified that $\bs_T$ is a fixpoint of $D_T$. Yet $D_T$
has a second, unintended fixpoint $\{\{P,Q\}\}$ which contains more knowledge
than $\bs_T$. This is a problem as it is this
unintended fixpoint that is obtained by
iterating $D_T$ starting with $\Worlds$. The reason for this mistake 
is that the
second, nonmonotonic inference rule applies in the initial stage
$\bs_0=\Worlds$ when $\neg K P$ holds. Later, when $P$ is derived,
the conclusion that $Q$ is a theorem continues to reproduce itself
through the third truth sayer rule. 

The problem above is that at each step and for each world $w$ an {\em
  assumption} is made of whether $w$ is possible or impossible.  Each
such an assumption might be right or wrong. These assumptions are
revised by iterated application of $D_T$. In the context of monotonic
inference rules, the only wrong assumptions that might be made during the
monotonic fixpoint construction starting in $\Worlds$ are that some world
is possible, while in fact it turns out to be impossible. But these
wrong assumptions can never lead to an erroneous application of an
inference rule: if a condition $K\varphi$ of  an inference rule
holds when $w$ is assumed to be possible, then it will still hold when
$w$ turns out to be impossible. But in the context of nonmonotonic
inference rules, an inference rule may fire due to an erroneous
assumption and its conclusion might be maintained through a circular
argument in all later iterations. In our scenario, it is the initial
assumption that worlds in which $P$ is false are possible that lead to
the assumption that worlds in which $Q$ is false are impossible, and
this assumption is later reproduced by a circular reasoning using
the third truth sayer proposition for $Q$.

The solution to this problem is very simple: \emph{never make any
unjustified assumption about the status of a world}. Start without any
assumption about the status of any worlds and only assign a specific
status when certain. We will elaborate this idea in two steps.
In the first step, we illustrate this idea for a simplification $T'$ 
of $T$, in which the third axiom $K Q \mim Q$ has been deleted.

\medskip
\noindent
1. Initially, no world is known to be possible or impossible. At
  this stage, the truth value of the unique modal literal $K P$ in
  $T'$ cannot be established. Yet, some things are clear. First,
  all worlds in which $P$ is false, that is, $\emptyset$ and $\{Q\}$, are
  certainly impossible since 
  they violate the first formula, $P$, of $T'$. 
  Second, the world $\{P, Q\}$ is definitely possible since no matter
  whether $P$ is a theorem or not, this world satisfies the two formulas
  of $T'$. 
  All this can be
  established without  making a single unsafe assumption. Thus, the only world 
  about which we are uncertain at this stage is the world $\{P\}$ in
  which $Q$ is false. Due to the second axiom, this world 
  is possible if $P$ is known and impossible otherwise. 

\smallskip
\noindent
2. In the next pass, we first use the knowledge that we gained in 
the previous step to re-evaluate the modal literal $K P$. In particular,
it can be seen that $P$ is true in all possible worlds and in the last
remaining world of unknown status, $\{P\}$. This suffices to establish 
that $P$ is a theorem, that is, that $K P$ is true. 

  With this newly gained information, we can establish the status of
  the last world and see that $\{P\}$ satisfies the two axioms of
  $T$. Hence this world is possible.

\medskip
The construction stops here. The next pass will not change anything,
and we obtain the possible-world set $\bs_{T'} = \{ \{P\}, \{P,Q\}\}$.
Now, let us add the third axiom $K Q \mim Q$ back and consider the full
theory $T$. 

\medskip
\noindent
1. The first step of the construction is identical to the one above
and determines the status for all worlds except $\{P\}$: $\{P,Q\}$ is
  possible, and $\emptyset$ and $\{Q\}$ are impossible.

\smallskip
\noindent
2. As before, in the second pass, $K P$ can be established to be true. 
  The second modal literal $K Q$ in $T$ cannot be established yet since 
  its truth depends on the status of the world $\{P\}$. The literal would
  be false if $\{P\}$ is possible, and true otherwise. Thus the truth of 
  the third axiom in $\{P\}$ is still undetermined. We are blocked here. 

\smallskip
\noindent
3. But there is a way out of the deadlock. So far, the methods to
  determine whether a world is possible or impossible were perfectly
  symmetrical. The solution lies in breaking this symmetry. In $T$, we
  have a truth-sayers axiom: it is consistent to assume that $Q$ is a
  theorem, and also to assume that $Q$ is not a theorem. In semantical
  terms, both assumptions on world $\{P\}$ are consistent: if this
  world is chosen possible, then $K Q$ is false and all axioms are
  satisfied in $\{P\}$; if it is chosen impossible, then $K Q$ is
  true. Since we want to interpret the modal operator as a theoremhood
  modality, it is clear what assumption to make: that $Q$ is
  not a theorem. We should make the assumption of {\em ignorance} and
  take it that the world is possible (and $Q$ is not a theorem). Thus,
  we obtain again the possible world model $\bs_T = \{\{P\}, \{P,Q\}\}$.

\medskip
From these two examples, we can extract the concepts necessary
to formalize the above informal reasoning processes. At each 
step, we have \emph{partial} information about the status of worlds that was
gained so far. This naturally formalizes as a 3-valued set of worlds. We call 
such a set a {\em partial possible-world set}. Formally, a partial 
possible-world set $\tbs$ is a function
\[
\tbs:\Worlds\ra\{\Tr,\Fa,\Un\},
\] 
where $\Worlds$ is the collection of all interpretations. Standard,
{\em total} 
possible-world sets can be viewed as special cases, where the only two
values in the range of the function are $\Tr$ and $\Fa$. In the
context of a partial possible world $\tbs$, we call a world $w$ {\em
  certainly possible} if $\tbs(w)=\Tr$ and {\em potentially possible}
if $\tbs(w)=\Tr$ or $\Un$. Likewise, we  call a world $w$ {\em
  certainly impossible} if $\tbs(w)=\Fa$ and {\em potentially impossible}
if $\tbs(w)=\Fa$ or $\Un$. If $\tbs(w)=\Un$, $w$ is potentially
possible and potentially impossible. We define $CP(\tbs)$ as the set
of certainly possible worlds of $\tbs$,  $PP(\tbs)$ as the set of
potentially possible worlds, and likewise, $CI(\tbs)$ and $PI(\tbs)$
as the sets of certainly impossible, respectively potentially
impossible worlds of $\tbs$.

At each inference step \wis{\tbs_i \ra \tbs_{i+1}}, we evaluated the
propositions of $T$ in one or more unknown worlds \wis{w}, given the
partial information available in \wis{\tbs_i}. When all propositions
of $T$ turned out to be true in $w$, $w$ was derived to be possible;
if some evaluated to false, $w$ was inferred to be impossible. To
capture this formally, we need a three-valued truth function to evaluate
theories in the context of a world $w$, the one we are examining, and 
a partial possible-world set $\tbs$. The value of this truth function on
a theory $T$, denoted as $\tf{T}{\tbs,w}$, is selected from
$\{\Tr,\Fa, \Un\}$. There are some obvious properties that this function 
should satisfy. 

\medskip
\noindent
1. The three-valued truth function should coincide with the
  standard (implicit) truth function for modal logic in total possible-world 
  sets.  In particular, when $\tbs$ is a total possible-world
  set, that is, $\tbs$ has no unknown worlds, then $\tf{T}{\tbs,w}$
  should be true precisely when $\tbs,w\models T$ (and false, otherwise).

\smallskip
\noindent
2. The three-valued truth function should be monotone with respect
  to the {\em precision} of the partial possible-world sets. A more
  precise partial possible-world set is one with fewer (with respect
  to inclusion) unknown worlds. 

\medskip
The intuition presented in (2) can be formalized as follows.
We define $\tbs \leqp \tbs'$ if
  $\tbs(w)\leqp\tbs'(w)$, where the latter (partial) order $\leqp$ 
  on truth values is the one generated by $\Un\leqp\Tr$ and $\Un\leqp\Fa$.

  A three-valued truth function $|T|^{\tbs,w}$ is monotone in $\tbs$ if
  $\tbs'\leqp \tbs''$ implies that $\tf{T}{\tbs',w} \leqp \tf{T}{\tbs'',w}$.
  In particular, if $|T|^{\tbs,w}$ is monotone in $\tbs$ and $\bs$ is a 
  total possible-world set such that $\tbs' \leqp \bs$, then 
  $\tf{T}{\tbs',w}=\Tr$ implies that $\bs,w\models T$, and $\tf{T}{\tbs',w}
  =\Fa$ implies that $\bs,w\not \models T$.

Designing such a three-valued truth function is routine, the problem is
that there is more than one sensible solution. One approach, originally 
proposed by \citet{DMT98}, extends Kleene's \citeyearpar{Kleene52}
three-valued truth evaluation 
to modal logic.

\begin{definition} 
For a formula $\vph$, world $w\in\Worlds$ and partial
possible-world set $\tbs$, we define \wis{\tf{\varphi}{\tbs,w}} using 
the standard Kleene truth evaluation rules of three-valued logic augmented
with one additional rule:
\[
|K\vph|^{\tbs,w} =
\left \{\begin{array}{ll}
\Fa & \mbox{ if } \tf{\vph}{\tbs,w'} =\Fa, \mbox{ for some $w'$ such
  that }\tbs(w')=\Tr \\
\Tr & \mbox{ if } \tf{\vph}{\tbs,w'} =\Tr, \mbox{ for all $w'$ such
  that $\tbs(w')=\Tr$ or $\Un$} \\
\Un & \mbox{ otherwise. }
\end{array} \right .
\]
For a theory $T$, we define $\tf{T}{\bs,w}$ in the standard way of
three-valued logic:
\[
|T|^{\bs,w} =
\left \{\begin{array}{ll}
\Fa & \mbox{ if } \tf{\vph}{\tbs,w} =\Fa, \mbox{ for some $\vph\in T$}\\
\Tr & \mbox{ if } \tf{\vph}{\tbs,w} =\Tr, \mbox{ for all $\vph\in T$}\\
\Un & \mbox{ otherwise. }
\end{array} \right .
\]
\end{definition}

\ignore{ I HAD THE HOPE TO BE ABLE TO SHOW THAT REITERS
  DEFINITION OF EXTENSIONS WITH ITS CHARACTERISTIC ASSYMMETRIC 
TREATMENT OF PREREQUISITES AND JUSTIFICATIONS, ARISES NATURALLY FROM 
OUR DEFINITIONS. IT IS NOT SO EASY.  
Hence, for an inference rule $\varphi$ of the form (\ref{eq:AELprop}):
\[ K \alpha_1 \land \dots \land K \alpha_n \land \neg K \beta_1 \land \dots 
\land \neg K \beta_m \rightarrow \gamma \]
we can see that $\ktf{\varphi}{\tbs,w}=\Tr\lor\Un$ if

$\bs$ is the least $\bs'$ such that $\bs'$ is the set of all worlds
$w$ that  satisfy all non-modal
formulas of $\Delta$ and for each inference rule 
\[ K \alpha_1 \land \dots \land K \alpha_n \land \neg K \beta_1 \land \dots 
\land \neg K \beta_m \rightarrow \gamma \]
such that $\bs'\models K\alpha_i$ and $\bs\not \models \neg K\beta_j$
then $w\models \gamma  $
}

To illustrate the use of this truth function, let us evaluate the
formula $K\varphi$, where $\varphi$ is objective, in the
context of a partial possible-world set $\tbs$ and an arbitrary world
$w$.  We have $\tf{K\varphi}{\tbs,w}=\Tr$ if $PP(\tbs) \models
K\varphi$, that is, if all potentially possible worlds satisfy
$\varphi$. Likewise, we have $\tf{K\varphi}{\tbs,w}=\Fa$ if $CP(\tbs)
\not \models K\varphi$, that is, at least one certainly possible world
violates $\varphi$. Let $\bs$ be a more precise total possible world
set; that is, $\tbs\leqp\bs$ or equivalently, $PP(\tbs) \supseteq \bs
\supseteq CP(\tbs)$.  Then, obviously, if $K\varphi$ holds true in
$\tbs$, the formula is true in $\bs$, and if $K\varphi$ is false in
$\tbs$ then it is false in $\bs$ as well. In general this truth
function is {\em conservative} (that is, $\leqp$-monotone) in the sense that
if a formula evaluates to true or false in some partial possible-world
set, then it has the same truth value in every more precise
possible-world set thus, in particular, in every total possible-world
set $\bs$ such that $\tbs \leqp \bs$.

It is easy to see (and it was proven formally by \citet{DMT03}) that
this truth function satisfies the two desiderata listed above. We also
note that this is not the only reasonable way in which the three-valued 
truth function can be defined. We will come back on this topic in 
Section \ref{sec:summary}.

We now review the framework of semantics of autoepistemic reasoning we 
introduced in our study of the relationship between the default logic 
of Reiter and the autoepistemic logic of Moore \citep{DMT03}. We listed 
these semantics in the previous section. 
All semantics in the framework require that a (partial) possible-world
model $\tbs$ of an autoepistemic theory be justified by some type of
an inference process:
\[ 
\tbs_0 \ra \tbs_1 \ra \dots \ra \tbs_n = \tbs. 
\]
At each step $i$, modal literals $K\varphi$ appearing in $T$ are
evaluated in $\tbs_i$. When such literals are derived to be true or
false, this might lead to further inferences in $\tbs_{i+1}$. Taking
the semantic point of view, we understand an inference here as a step
in which some worlds of undetermined status are derived to be possible 
and some others are derived impossible.

Dialects of autoepistemic logic, and so of default logic, too, differ 
from each other in the nature of the derivation step $\tbs_i \ra \tbs_{i+1}$,
and in initial assumptions $\tbs_0$ they make. Some dialects make no initial 
assumptions at all; in some others making certain initial ``guesses'' 
is allowed. In this way, we obtain autoepistemic logics of different 
degrees of {\em groundedness}. In the following sections, we describe 
inference processes underlying each of the four semantics
in the framework described in Section \ref{sec:moredefaultsvsael}.

Finally, we link the above concepts with the algebraic lattice
theoretic concepts sketched in the previous section and used in the
semantic framework of \citet{DMT03}. There, the different semantics of an
autoepistemic theory $T$ emerged as different types of fixpoints of a
$\leqp$-monotone operator $\D_T$ on the bilattice consisting of
arbitrary pairs $(\bs,\bs')$ of possible-world sets. The partial
possible-world sets $\tbs$ correspond to the {\em consistent} pairs
$(PP(\tbs),CP(\tbs))$ in this bilattice; a pair $(\bs,\bs')$ is {\em
  consistent} if $\bs\supseteq \bs'$, that is, certainly possible worlds
are potentially possible. Inconsistent pairs give rise to 
possible-world sets that in addition to truth values $\Tr$, $\Fa$
and $\Un$ require the fourth one $\In$ for ``inconsistency''.
The Kleene truth function defined above can be
extended easily to a four-valued truth function on the full
bilattice. 

The operator $\D_T$ on that bilattice was then defined as follows:
\[
\D_T(\tbs)= \tbs', \mbox{ if for every $w\in\Worlds$, } 
\tbs'(w)=\tf{T}{\tbs,w}.
\] 
We observe that this operator maps partial possible-world sets into
partial possible-world sets and that it coincides with Moore's
derivation operator $D_T$ when applied on total possible-world sets. 

In the sequel, we will often represent a partial possible-world set
$\tbs$ in its bilattice representation, as the pair
$(PP(\tbs),CP(\tbs))$ of respectively potentially possible and
certainly possible worlds. For example, the least precise partial
possible-world set $\bot_p$ for $\Sigma=\{P,Q\}$ will be written as
$(\verz{\emptyset,\{P\}, \{Q\},\{P,Q\}},\emptyset)$: all worlds are
potentially possible; no world is certainly possible.

We will now discuss the four semantics discussed above that define
different dialects of autoepistemic reasoning.

\subsection{The Kripke-Kleene semantics}

This semantics is a direct formalization of the discussion above.
We are given a finite modal theory $T$ (we adopt the assumption of finiteness
to simplify presentation, but it can be omitted). A \emph{Kripke-Kleene 
inference process} is a sequence 
\[
\tbs_0 \ra \dots \ra \tbs_n
\]
of partial possible-world sets such that:

\medskip
\noindent
1. $\tbs_0$ is the totally unknown partial possible-world set. That
  is, for every $w\in\Worlds$, $\tbs_0(w)=\Un$. We denote this partial
  possible-world set by $\bot_p$. This choice of the starting point
  indicates that
  Kripke-Kleene inference process does not make any initial
  assumptions.

\smallskip
\noindent
2. For each $i=0,\ldots,n-1$, there is a set of worlds $U$ such
  that for every $w\in U$, $\tbs_{i}(w) =\Un$,
  $\tf{T}{\tbs_i,w}\not=\Un$ and $\tbs_{i+1}(w)=\tf{T}{\tbs_i,w}$, and
  for every $w\notin U$, $\tbs_{i}(w)=\tbs_{i+1}(w)$. Thus, in each
  step of the derivation the status of the worlds that are certainly
  possible and certainly impossible does not change. All that can change is the
  status of some worlds of unknown status (worlds, that are
  potentially possible and potentially impossible). This set is
  denoted by $U$ above. It is not necessary that $U$ contains all
  worlds that are unknown in $\tbs_{i}$. In the derivation, worlds
  in $U$ become certainly possible or certainly impossible, depending on how the
  theory $T$ evaluates in them. If for such a potentially possible
  world $w\in U$, $\tf{T}{\tbs_i,w}=\Tr$, $w$ becomes certainly
  possible. If $\tf{T}{\tbs_i,w}=\Fa$, $w$ becomes certainly
  impossible. Otherwise, the status of $w$ does not change. As such a
  derivation starts from the least precise, hence assumption-free,
  partial possible-world set $\bot_p$, all these derivations are
  assumption-free.

\smallskip
\noindent
3. The halting condition: no more inferences can be made once we reach
the state $\tbs_n$. Here this means  that for each unknown $w\in\Worlds$,
  $\tf{T}{\tbs_n,w}=\Un$.  The process terminates.
 
\medskip
This precise definition formalizes and generalizes the informal 
construction we presented in the previous section. When applied to the 
theory we considered there,
\[
T' = \verz{ P,  \neg K P \mim Q },
\]
one Kripke-Kleene inference process that might be produced is (we 
represent here worlds, or interpretations, as sets of atoms they
satisfy, and partial possible-world sets $\tbs$ as pairs $(PP(\tbs),CP(\tbs))$):
\[ 
\begin{array}{lll}
\bot_p 
  & \ra \tbs_1=(\verz{\emptyset, \{P\},\{P,Q\}}, \emptyset) 
   & \quad\mbox{$\{Q\}$ certainly impossible}\\
  & \ra \tbs_2=(\verz{\{P\},\{P,Q\}}, \emptyset) 
   & \quad\mbox{$\emptyset$  certainly impossible}\\ 
  & \ra \tbs_3=(\verz{\{P\},\{P,Q\}},\verz{\{P,Q\}})  &
  \quad\mbox{$\{P,Q\}$ 
     certainly possible}\\
  & \ra \tbs_4=(\verz{\{P\},\{P,Q\}}, \verz{\{P\},\{P,Q\}}) & \quad\mbox{$\{P\}$ 
 certainly possible}.
\end{array}
\]
The first derivation can be made since
$\tf{P\land(\neg K P \ra P)}{\bot_p,w}=\Fa$, for $w=\{Q\}$ (in fact, 
for every $w$, in which $P$ is false). The second derivation is justified 
similarly as the first one. The third derivation follows as
$\tf{P\land(\neg K P\ra Q)}{\tbs_2,w}=\Tr$, for $w=\{P,Q\}$, and the 
forth one as $\tf{P\land(\neg K P \ra Q)}{\tbs_3,w}=\Tr$, for $w=\{P\}$. Let 
us explain one more detail of the last of these claims. Here, $\tf{P}{\tbs_3,w}=
\Tr$ holds because $P$ holds in $w=\{P\}$. Moreover, $\tf{KP}{\tbs_3,w}=
\Tr$ as $P$ holds in every world that is potentially possible in $\tbs_3$.
Thus, $\tf{\neg KP}{\tbs_3,w}=\Fa$ and so indeed, 
$\tf{\neg KP\ra Q}{\tbs_3,w}= \Tr$.

The shortest derivation sequence that corresponds exactly to the
informal construction of the previous section is: 
\[ 
\bot_p \ra (\{\{P\},\{P,Q\}\},\{\{P,Q\}\}) \ra 
(\{\{P\},\{P,Q\}\},\{\{P\},\{P,Q\}\}). 
\]
The fact that there may be multiple Kripke-Kleene inferences processes
is not a problem as all of them end in the same partial possible-world.

\begin{proposition}
  For every modal theory $T$, all Kripke-Kleene inference processes 
  converge to the same partial possible-world set, which is the 
  $\leqp$-least fixpoint of the operator $\D_T$. 
\end{proposition}

We call this special partial possible-world set the \emph{Kripke-Kleene 
extension} of the modal theory $T$.

While the Kripke-Kleene construction is an intuitively sound construction,
it has an obvious disadvantage: in general, its terminating partial belief
state may not match the intended belief state even if $T$ consists of
``monotonic'' inference rules (no negated modal atoms in the antecedents
of formulas of the form (\ref{eq:AELprop})). An example where this happens
is the truth sayer theory:
\[
T = \verz{ K P \mim P }.
\]
It consists of a single monotonic inference rule, and its
intended total possible-world set is $\verz{\{\emptyset,\{P\}}$, which
in the current $(PP,CP)$ notation corresponds to
\[
(\verz{\{\emptyset,\{P\}},\verz{\{\emptyset,\{P\}}).
\]
However, the one and only
Kripke-Kleene construction is
\[
\bot_p \ra (\verz{\emptyset, \{P\}}, \verz{\{P\}}). 
\] 
Then the construction halts. No more Kripke-Kleene inferences on the 
status of worlds can be made and the intended possible-world set is not
reached.

We conclude with a historical note. The name Kripke-Kleene semantics 
was used for the first time in the context of the semantics of logic 
programs by \citet{Fitting85}. Fitting built on ideas in an earlier work
by \citet{Kleene52}, and on Kripke's \citeyearpar{Kripke75} theory of truth,
where Kripke discussed how to handle the liar paradox. 

\subsection{Moore's autoepistemic logic}

Moore's autoepistemic logic has a simple formalization in our
framework. A possible-world set $\bs$ is an autoepistemic
expansion of $T$ if there is a \emph{one-step} derivation for it:
\[
\tbs_0 \ra \tbs_1,
\]
where $\tbs_0 = \tbs_1 = \bs$. Clearly, here we allow the inference
process to make initial assumptions. Moreover, in the derivation step
$\tbs_0 \ra \tbs_1$ we simply verify that we made no incorrect 
assumptions and that no additional inferences can be drawn. The 
inference (more accurately here, the verification) process works as
follows:

\medskip
\noindent
1. A world $w$ is derived to be possible if $\tbs_0,w\models T$.

\smallskip
\noindent
2. A world $w$ is derived to be impossible if $\tbs_0,w\not\models T$. 

\medskip
Thus, formally, $\tbs_1 = \{ w \st \tbs_0,w\models T\} = D_T(\tbs_0)$.
Consequently, the limits of this derivation process are indeed precisely 
the fixpoints of the Moore's operator $D_T$ (we stress that we talk here 
only about total possible-world sets).

Since $\D_T$  coincides with $D_T$ on total possible-world sets,
all autoepistemic expansions are fixpoints of $\D_T$. Thus, we have the
following result.
\begin{proposition} 
  The Kripke-Kleene extension is less precise than any other
  autoepistemic expansion of $T$. If the Kripke-Kleene extension is 
  total, then it
  is the unique autoepistemic expansion of $T$.
\end{proposition}

The weakness of Moore's logic from the point of view of modeling the
autotheoremhood view has been argued above. In Section \ref{sec:other}, 
we will discuss another interpretation of autoepistemic logic in which his
semantics may be more adequate.

\subsection{The well-founded knowledge derivation}

The problem with the Kripke-Kleene derivation is that it treats
ignorance and knowledge in the same way. Ignorance is reflected 
by the presence of possible worlds. Knowledge is reflected by the 
presence of impossible worlds. In the Kripke-Kleene derivation, 
both possible and impossible worlds are derived in a symmetric 
way, by evaluating the theory $T$ in the context of a world $w$, given 
the partial knowledge $\tbs$. 

What we would like to do is to impose ignorance as a default. That a world is 
possible should not have to be derived. A world should be possible 
{\em unless} we can show that it is impossible. In other words, we need
to impose a principle of {\em maximizing ignorance}, or equivalently,
{\em minimizing knowledge}. Under such a principle, it is obvious that
the possible-world set $\{\{P\}\}$ cannot be a model of the truth sayer
theory $T = \verz{ KP \mim P }$. It does not minimize knowledge while the 
other candidate for a model, the possible-world set $\{\emptyset,
\{P\}\}$, does.

To refine the Kripke-Kleene construction of knowledge, we need an
additional derivation step that allows us to introduce the assumption 
of ignorance. Intuitively, in such a derivation step, we consider 
a set \wis{U} of unknown worlds, which are turned into certainly 
possible worlds to maximize ignorance.  

Formally, a \emph{well-founded inference process} is a derivation 
process $\tbs_0 \ra \dots \ra \tbs_n$ that satisfies the same 
conditions as a Kripke-Kleene inference process except that some 
derivation steps $\tbs_i \ra \tbs_{i+1}$ may also be justified as 
follows (by the \emph{maximize-ignorance} principle):
\begin{itemize}
\item[MI:] There is a set $U$ of worlds such that $\tbs_{i+1}(w)=\tbs_i(w)$
  for all $w\not\in U$ and for all $w\in U$, $\tbs_i(w)=\Un, 
  \tbs_{i+1}(w)=\Tr$ and $\tf{T}{\tbs_{i+1},w}=\Tr$.
\end{itemize}
In other words, in such a step we pick a set $U$ of unknown words,
assume that they are certainly possible, and verify that this
assumption was justified, that is, under the increased level of
ignorance, all of them turn out to be certainly possible.  To put it
yet differently, we select a set $U$ of unknown worlds, for which it
is consistent to assume that they are certainly possible, and we turn
them into certainly possible worlds (increasing our ignorance). By 
analogy with the notion of an unfounded set of atoms \citep{Vangelder91},
we call the set of worlds $U$, with respect to which the
maximize-ignorance principle applies at the partial belief state
$\tbs_i$, an \emph{unfounded} set for $\tbs_i$. 

We also note that the halting condition of a well-founded inference
process is stronger than that for a Kripke-Kleene process. This means
that for each unknown world $w$ of $\tbs_n$, $\tf{T}{\tbs_n,w}=\Un$
and in addition, $\tbs_n$ does not allow a MI inference step, that is, it
has no non-empty unfounded set.

There are two properties of well-founded inference processes that are
worth noting.

\begin{proposition}
All well-founded inference processes converge to the same (partial) 
possible-world set. 
\end{proposition}

This property gives rise to the {\em well-founded extension} of the
modal theory \wis{T} defined as the limit of \emph{any} well-founded
inference process.  This limit can be shown to coincide
with the well-founded fixpoint of $\D_T$, that is, the \wis{\leqp}-least
fixpoint of the operator $\wis{\curly{S}_{\D_T}}$ defined in the
previous section.

Another important property concerns theories with no positive 
occurrences of the modal operator (for instance, theories consisting
of formulas (\ref{eq:AELprop}) with no modal literals $\neg K\beta_j$ in
the antecedent). 

\begin{proposition}
If $T$ contains only negative occurrences of the modal operator, then
the well-founded extension is the $\leq_k$-least fixpoint of $D_T$.
\end{proposition}

This property shows that the well-founded extension semantics has all
key properties of the desired semantics of sets of ``monotonic inference
rules.'' Let us revisit the truth sayer theory:
\[
T = \verz{ KP \mim P }.
\]
The  Kripke-Kleene construction is 
\[ 
\bot_p \ra (\verz{\emptyset,\{P\}}, \verz{\{P\}}).
\]
The inference that $\{P\}$ is possible is also sanctioned under the
rules of the well-founded inference process. However, while there is
no Kripke-Kleene derivation that applies now, the maximize-ignorance
principle does apply and the well-founded inference process can
continue. Namely, in the belief state given by
$(\verz{\emptyset,\{P\}}, \verz{\{P\}})$, there is one world of
unknown status (neither certainly impossible, nor certainly possible):
$\emptyset$.  Taking $U = \{\emptyset\}$ and applying the
maximize-ignorance principle to $U$, we see that the well-founded
inference process extends and yields $(\verz{\emptyset,\{P\}},
\verz{\emptyset,\{P\}})$. This possible-world set is total and so,
necessarily, the limit of the process. Thus, this (total) possible-world 
set $\verz{\emptyset,\{P\}}$ is the well-founded
extension of the theory $\{ KP \mim P\}$.

The well-founded extension is total not only for monotonic theories.
For instance, let us consider the theory:
\[
T = \verz{ KP \leftrightarrow Q } \mbox{\ \ \ or equivalently, }\ 
\verz{KP\ra Q, \neg KP \ra \neg Q}.
\]
Intuitively, there is nothing known about $P$, hence $Q$ should be 
false. The unique Kripke-Kleene inference process ends
where it starts, that is, with $\bot_p$. Indeed, when $KP$ is 
unknown, no certainly possible or certainly impossible worlds can be 
derived. However, the 
possible-world set $U=\verz{\emptyset,\{P\}}$ is unfounded with respect 
to $\bot_p$. Indeed, if both worlds are assumed possible,  $KP$ evaluates
to false, and both worlds satisfy $T$. Thus, in the well-founded derivation 
we can establish that and then, in the next two steps, we can
derive the impossibility of the two remaining unknown worlds, first of 
$\{Q\}$ and then of $\{P,Q\}$. This yields the following well-founded 
inference process: 
\[  
\begin{array}{ll}
   \bot_p & \ra \tbs_1=(\verz{\emptyset,\{P\},\{Q\},\{P,Q\}}, 
                        \verz{\emptyset,\{P\}})\\
    & \ra \tbs_2=(\verz{\emptyset,\{P\},\{P,Q\}}, 
                        \verz{\emptyset,\{P\}})\\
   & \ra \tbs_3=(\verz{\emptyset,\{P\}}, \verz{\emptyset,\{P\}}).
\end{array}
\]

In other cases, the well-founded extension is a partial
possible-world set. An 
example is the theory: 

\[
\verz{\neg KP \ra Q, \neg KQ\ra P\}}.
\]
In this case, there is only one well-founded inference process, which
derives that \wis{\{P,Q\}} is a certainly possible world and derives
no certainly impossible worlds. That is, the
well-founded extension is: $(\verz{\emptyset,\{P\},\{Q\},\{P,Q\}}, 
\verz{\{P,Q\}})$. 

\subsection{Stable possible-world sets}

We recall that a partial possible-world set $\tbs$ corresponds to the
pair of total possible-worlds sets: $(PP(\tbs),CP(\tbs))$, where 
$PP(\tbs)$ is the set of potentially possible worlds and $CP(\tbs)$ is 
the set of certainly possible worlds. 

We now define a {\em stable derivation} for a possible-world set
\wis{\bs} as a sequence of partial belief states of the form:
\[
(\Worlds,\bs) \ra (PP_1,\bs) \ra  \dots \ra (PP_{n-1},\bs) \ra (PP_n,\bs),
\]
where:

\medskip
\noindent
1. $PP_n=\bs$

\smallskip
\noindent
2. For every $i=0,\ldots,n-1$, and for every $w\in PP_i\setminus
  PP_{i+1}$, $\tf{T}{(PP_i,\bs),w}=\Fa$. That is, some worlds $w$ in which $T$
  is false with respect to $\tbs_i = (PP_i,\bs)$ become certainly
  impossible and are removed from $PP_i$ to form $PP_{i+1}$.

\smallskip
\noindent
3. Halting condition: for every $w\in PP_n$, $\tf{T}{(PP_n,\bs),w}=\Tr$ 
or $\Un$.

\medskip
If a total belief set \wis{\bs} has a stable derivation then we call 
$\bs$ a {\em stable extension}. This concept captures the idea of the 
Reiter's extension of a default theory.

We recall that an inference rule (\ref{eq:AELprop}) evaluates to false in
world $w$ with respect to $(PP_i,\bs)$ if $w\not\models \gamma$,
$PP_i\models K\alpha_i$, for all $i$, $0\leq i \leq n$, and $\bs\not\models
K\beta_j$, for all $j$, $0\leq j \leq m$. We see here an asymmetric
treatment of prerequisites $\alpha_i$ and justifications $\beta_j$
which are evaluated in two different possible world
sets. The same feature shows up, not coincidentally, in Reiter's
definition of extension of a default theory.

The intuition underlying a stable derivation comes from a different
implementation of the idea that ignorance does not need to be 
justified and that only knowledge must be justified. In a partial 
possible-world set $\tbs$, the component sets $PP(\tbs)$ and $CP(\tbs)$
have different roles. Since $PP(\tbs)$ determines the certainly impossible 
worlds, this is the possible-world set that determines what is definitely
known. On the other hand the set $CP(\tbs)$ of certainly possible worlds
determines what is definitely not known by $\tbs$.

A stable derivation for $\bs$ is a justification
for each impossible world of $\bs$ (each world is initially potentially
possible but eventually determined not to be in $\bs$, that is,
determined impossible in $\bs$). The key point is that this
justification may use the assumption of the ignorance in $\bs$. By
fixing $CP(\tbs_i)$ to be $\bs$, it takes the ignorance in $\bs$ for
granted. What is justified in a stable inference process is the
impossible worlds of $\bs$, not the possible worlds.

We saw above that the theory 
\[
\verz{\neg KP \ra Q, \neg KQ\ra P}
\] 
has a partial well-founded extension. It turns out that it has two stable
extensions $\verz{\{P\},\{P,Q\}}$ and $\verz{\{Q\},\{P,Q\}}$. For 
instance, the following stable derivation reconstructs
$\bs=\verz{\{P\},\{P,Q\}}$. Note that in any partial possible-world set
$(\cdot,\bs)$ (that is, where the worlds of $\bs$ are certainly possible),
$KQ$ evaluates to false. In all such cases, $T$ evaluates to false in any
world in which $P$ is false. Hence we have the following very short stable 
derivation:

\[  \begin{array}{lll}
   (\Worlds,\bs)  & \ra  (\bs,\bs).
  \end{array}\]

\ignore{
\begin{proposition}
All terminal stable inference processes for $\bs$ converge to the same
limit. 
\end{proposition}
}

We now have two key results. The first one links up well-founded and
stable extensions. 

\begin{proposition}  If the well-founded extension is a total
  possible-world set, it is the unique stable extension.
\end{proposition}

The second result shows that indeed, the Konolige's translation works
if the semantics of default logic of Reiter and the autoepistemic logic
of Moore are correctly aligned. Here we state the result for the most 
important
case of default extensions and stable extensions, but it extends, as we noted
earlier, to all semantics we considered.

\begin{proposition} For every default theory $\Delta$, $\bs$
  is an extension of $\Delta$ if and only $\bs$ is a stable extension of
  $\Kon(\Delta)$. 
\end{proposition}

\subsection{Discussion}
\label{sec:summary}

We have obtained a framework with four different semantics. This
framework is parameterized by the truth function. We have
concentrated on the Kleene truth function but other viable choices
exist. One is super-valuation \citep{vanFraassen66} which defines
$\tf{T}{\tbs,w}$ in terms of the evaluation of $T$ in all possible
world sets $\bs \geq_P \tbs$ approximated by $\tbs$. In particular,
\[ 
\tf{T}{\tbs,w}=Min_{\leq_p}\{ \tf{T}{\bs,w} \st \tbs \leq_p \bs\}.
\]
\ignore{
\begin{itemize}
\item  \wis{\tf{T}{\tbs,w}=\Tr} if for every possi\wis{\bs} approximated by
  \wis{\tbs}, $\bs,w\models T$.
\item \wis{\stf{T}{\tbs,w}=\Fa} if for every \wis{\bs} approximated by
  \wis{\tbs}, $\bs,w\not \models T$.
\item \wis{\stf{T}{\tbs,w}=\Un} otherwise. 
\end{itemize}
}
In this way we obtain another instance of the framework, the
family of \emph{ultimate} semantics \citep{dmt04}.  For many theories, 
the corresponding semantics of the two families coincide but ultimate 
semantics are sometimes more
precise. An example is the theory $\{ K P \lor \neg K P \mim P
\}$. It's Kripke-Kleene and well-founded extension is the partial
possible world set $(\{\emptyset,\{P\}\}, \{\{P\}\})$ and there are no
stable extensions. But the premise $K P \lor \neg K P$ is a
propositional tautology, making $\tf{T}{\bot_p,w}$ true if $w\models P$
and false otherwise. As a consequence, the ultimate Kripke-Kleene,
well-founded and unique stable extension is $\{\{P\}\}$.

\ignore{
We have obtained a framework with four different semantics, parameterized 
further by the truth function. We concentrated on the Kleene truth 
function in this article. But another viable choice is to use 
super-valuations \cite{vanFraassen66}, which leads to the family of 
\emph{ultimate} semantics \cite{dmt04}. Ultimate semantics have quite
attractive properties. For instance, the corresponding well-founded 
semantics is more precise than the one induced by the Kleene truth 
function. }

For a scientist interested in the formal study of the informal semantics 
of a certain type of (informal) propositions this diversity is troubling.
Indeed, what is then the nature of autoepistemic reasoning, and which
of the semantics that we defined and that can be defined by means of other
truth functions is the ``correct'' one?  It is necessary to bring
some order to this diversity. 

In the autotheoremhood view, the formal semantics should capture the
information content of an autoepistemic theory $T$ that contains
propositions referring to $T$'s own information content; the semantics
should determine whether a world is possible or impossible, or
equivalently, whether a formula is or is not entailed by $T$.
As we saw, Moore's semantics of expansions and the Kripke-Kleene extension
semantics are arguably less suited in the case of monotonic inference
rules with cyclic dependencies (cf. the truth sayer theory). This leaves 
us with four contenders only: the well-founded and
the stable extension semantics and their ultimate versions. All employ
a technique to maximize ignorance and correctly handle
autoepistemic theories with monotonic inference rules.  Which of these
semantics is to be preferred?

Let us first consider the choice of the truth function. The semantics
based on the Kleene truth function and the ones induced by
super-valuation make different trade-offs: the higher precision of the
ultimate semantics, which is good, comes at the price of higher
complexity of reasoning, which is bad \citep{dmt04}. When there is a
trade-off between different desired characteristics, there is per
definition no {\em best} solution. Yet, when looking closer, the
question of the choice between these two truth functions turns out to 
be largely {\em academic} and without much practical relevance. There are
classes of autoepistemic theories for which the Kleene and the
super-valuation truth functions coincide, and hence, so do the
semantics they induce. \citet[Proposition 6.14]{dmt04} provide an example 
of such a class.
Even more importantly, the semantics
induced by Kleene's truth function and by super-valuation differ only
when case-based reasoning on modal literals is necessary to make
certain inferences. Except for our own artificial examples
introduced to illustrate the formal difference between
both semantics \citep{dmt04}, we are not aware of any reasonable autoepistemic or
default theory in the literature where such reasoning would be
necessary. They may exist, but if they do, they will constitute an
insignificant fringe.  The take-home message here is that in all
practical applications that we are aware of, the Kleene truth function
suffices and there is no need to pay for the increased complexity of
super-valuation. This limits the number of semantics still in the
running to only two. Of the remaining two, the most faithful
formalization of the autotheoremhood view seems to be the well-founded
extension semantics. As we view a theory as a set of inference
rules, the construction of the well-founded extension formalizes
the process of the application of the inference rules more directly 
than the construction of the stable extension semantics.

Nevertheless, there are some commonsense arguments for not
overemphasizing the differences between these semantics. First, we
should keep in mind that theories of interest are those that are
developed by human experts, and hence, are meaningful to them. What
are the meaningful theories in the autotheoremhood?  Not every
syntactically correct modal theory makes sense in this view.
``Paradoxical'' theories such as the liar theory $T_{\mathit{liar}}$ can simply
not be ascribed an information content in a consistent manner and are
not a sensible theory in the autotheoremhood view.  For theories $T$
viewed as sets of inference rules, the inference process associated
with the theory should be able to determine the possibility of each
world and hence, for each proposition, whether it is a theorem or not
of $T$. In particular, this is the case when the well-founded
extension is total.  We view theories with theorems that are subject
to ambiguity and speculation with suspicion. And so, methodologies
based on the autotheoremhood view will naturally tend to produce
theories with a total well-founded extension. From a practical point
of view, the presence of a unique, constructible state of belief for
an autoepistemic theory is a great advantage. For instance, unless the
polynomial hierarchy collapses, for such theories the task to
construct the well-founded extension and so, also the unique stable
expansion, is easier than that of computing a stable expansion of an
arbitrary theory or to determine that none exists. Further, for such
theories, skeptical and credulous reasoning (with respect to stable
extension) coincide and are easier, again assuming that the polynomial
hierarchy does not collapse, than they are in the general case.

For all the reasons above, a human expert using autoepistemic logic in
the autotheoremhood view, will be naturally inclined to build an
autoepistemic knowledge base with a well-founded extension that is
total.  When the well-founded semantics induced by the Kleene truth
function is total, the four semantics --- the two stable semantics and
the two well-founded semantics --- coincide!  It is so, in particular
for the class of theories built of formulas (\ref{eq:AELprop}) with no
recursion through negated modal literals (the so-called
\emph{stratified} theories \citep{ge87}). Hence, such a methodology
could be enforced by imposing syntactical conditions. 

All these arguments notwithstanding, the fact is that many default theories 
discussed in the literature or arising in practical settings do not 
have a unique well-founded extension and that the stable and well-founded 
extension
semantics do not coincide\footnote{Some researchers believe that
  multiple extensions are {\em needed} for reasoning in the context of
  incomplete knowledge. Our point of view is different. The essence of
  incomplete knowledge is that different states of affairs are
  possible. Therefore, the natural --- and standard --- representation
  of a belief state with incomplete knowledge is by one possible-world
  set with multiple possible worlds, and not by multiple
  possible-world sets, which to us would reflect the state of mind of
  an agent that does not know what to believe.}.  We have seen it
above in the Nixon Diamond example. More generally, it is the case
whenever the theory includes conflicting defaults and no guidance on
how to resolve conflicts.  Such conflicts may arise inadvertently for
the programmer, in which case a good strategy seems to be to analyze
the conflicts (potentially by studying the stable extensions) and to
refine the theory by building in conflict-resolution in the conditions
of default rules. Otherwise, when conflicts are a deliberate decision
of the programmer who indeed does not want to offer rules to resolve
conflicts, all we can do is to accept each of the multiple stable
extensions as a possible model of the theory and also accept that none
of them is in any way preferred to others.

In conclusion, rather than pronouncing a strong preference for the
well-founded extension over stable extensions or vice versa, what we
want to point out is the attractive features of theories for which
these two semantics coincide, and advantages of methodologies that
lead to such theories.

\section{Autoepistemic Logics in a Broader Landscape} 
\label{sec:other}

In this section, we use the newly gained insights on the
nature of autoepistemic reasoning to clarify certain aspects of autoepistemic
logic
and its position  in the spectrum of logics, in particular in
the families of logics of nonmonotonic reasoning and classical modal
logics.

A good start for this discussion is Moore's ``second'' view on
autoepistemic logic. Later in his paper, when developing the expansion
semantics, Moore rephrased his views on autoepistemic reasoning in
terms of the background concept of an {\em autoepistemic agent}. Such
an agent is assumed to be ideally rational  and have the powers of perfect
introspection. An autoepistemic theory $T$ is viewed as a set of
propositions that are known by this agent. Modal literals $K\varphi$ in $T$
now mean {\em ``I (that is, the agent) know $\varphi$''}. The most
important assumption, the one on which this informal view of
autoepistemic logic largely rests, is that the agent's theory $T$
represents {\em all the agent knows} \citep{le90} or, in Moore's
terminology, what the agent knows is {\em grounded} in the theory. We
will call this implicit assumption the \AIKA.

Without the \AIKA, the theory $T$ would be just a list of believed
introspective propositions. The state of belief of the agent might
then correspond to any possible-world set $\bs$ such that $\bs\models
K\varphi$, for each $\varphi\in T$ (where $B \models K\varphi$ if for all $w\in B, B,w\models \varphi$). 
But in many such possible-world sets $\bs$, the agent would know much
more than what can be derived from $T$. In this setting, nonmonotonic
inference rules such as $K A(x) \land \neg K \neg B(x) \mim B(x)$
would not be useful for default reasoning since conclusions drawn from
them would not be derived from the information given in $T$.  So the
problem is to model the \AIKA in the semantics. Moore implemented this
condition by imposing that for any model $\bs$, if
 $\bs,w\models T$, then $w$ is possible according to $\bs$,
i.e., $w\in\bs$.
Combining both conditions, models that satisfy the \AIKA are fixpoints
of $D_T$, that is Moore's expansions.

Moore's expansion semantics does not violate the
assumptions underlying the autoepistemic agent view.  Expansions do
correspond to belief states of an ideally rational, fully
introspective agent that believes all axioms in $T$ and, in a sense,
does not believe more that what he can {\em justify} from $T$. But the
same can be said for the autotheoremhood view as implemented in the
well-founded and stable extension semantics. We may identify the
theory with what the agent knows, and the theoremhood operator with
the agent's epistemic operator $K$, and see the well-founded extension
(if it is total) or stable extensions as representing belief states 
of an agent that can be {\em justified} from $T$. 

As we stated in the previous section, Moore's expansion semantics does
not formalize the autotheoremhood view, but it formalizes a dialect of
autoepistemic reasoning, based on an autoepistemic agent that accepts
states of belief with a weaker notion of justification, allowing for
{\em self-supporting} states of belief. While not appropriate for
modeling default reasoning, the semantics may work well in other
domains.  Indeed, humans sometimes do hold self-supporting
beliefs. For example, self-confidence, or lack of self-confidence
often are to some extend self-supported. Believing in one's own
qualities makes one perform better. And a good performance supports
self-confidence (and self-esteem).  Applied to a scientist, this loop
might by represented by the theory consisting of the following formulas:

\begin{quote}
$K(ICanSolveHardProblems) \mim Happy$\\ 
$Happy \mim Relaxed$\\
$Relaxed \mim ICanSolveHardProblems$.
\end{quote}

Along similar lines, the placebo effect is a medically well-researched 
fact often attributed to self-supporting beliefs. The self-supporting
aspect underlying the placebo-effect can be described by the theory
consisting of the rules:

\begin{quote}
$K(IGetBetter) \mim Optimistic$\\
$Optimistic \mim IGetBetter$.
\end{quote}

Taking a placebo just flips the patients into the belief that they
are getting better. In this form of autoepistemic reasoning of an agent,
self-supporting beliefs are justified and Moore's expansion semantics,
difficult to reconcile with the notion of derivation and theorem, may
be suitable. 

There are yet other instances of the \AIKA in the autoepistemic agent
view. For example, let us consider the theory $T = \{ KP \}.$ In the
autotheoremhood view, this theory is clearly inconsistent, for there is no
way this theory can prove $P$.  The situation is not so
clear-cut in the agent view. We see no obvious argument why the agent
could not be in a state of belief in which he believes $P$ and its
consequences and nothing more than that. In fact, the logic of {\em
  minimal knowledge} \citep{HalpernM84} introduced as a variant of
autoepistemic logic accepts this state of belief for $T$.

What our discussion shows is that the \AIKA in Moore's autoepistemic
agent view is a rather vague intuition, which can be worked out in
more than one way, yielding different formalizations and different
dialects. It may explain why Moore built a semantics that did not satisfy
his own first intuitions (inference rules) and why \citet{Halpern97} 
could build
several formalizations for the intuitions expressed by Reiter and
Moore.  In contrast, the autotheoremhood view
eliminates the agent from the picture and hence, the difficult tasks
to specify carefully the key concepts such as ideal rationality,
perfect introspection and, most of all, the \AIKA. Instead, it builds
on more solid concepts of inference rules, theoremhood and entailment
which yields a more precise intuition.

\section{Conclusions}

We presented here an analysis of informal foundations of autoepistemic
reasoning. We showed that there is principled way to arrive at all 
major semantics of logics of autoepistemic reasoning taking as the
point of departure the autotheoremhood view of a theory. We see the main 
contributions of our work as follows.

First, extending Moore's arguments we clarified the different nature
of defaults and autoepistemic propositions. Looking back at Reiter's
intuitions, we now see that, just as Moore had claimed about McDermott
and Doyle, also Reiter built an autoepistemic logic and not a logic of
defaults. We showed that some long-standing problems with default
logic can be traced back to pitfalls of using the autoepistemic
propositions to encode defaults. On the other hand, we also showed that
once we focus theories understood as consisting of autoepistemic 
propositions and adopt the autotheoremhood perspective, we are led naturally
to the Kripke-Kleene semantics, the semantics of expansions by Moore, 
the well-founded semantics and the semantics of extensions by Reiter.

Second, we analyzed what can be seen as the center of
autoepistemic logic: the \AIKA. We showed that this rather fuzzy
notion leads to multiple perspectives on autoepistemic reasoning and 
to multiple dialects of the autoepistemic language, induced by different
notions of what can be derived from (or is {\em grounded in}) a theory. 
One particularly useful informal perspective on autoepistemic logic 
goes back to Moore's truly insightful view of autoepistemic rules as 
inference rules. This view, which we called the autotheoremhood view,
was the main focus of our discussion. In this view, theories ``contain'' 
their own entailment operator and ``I'' in the \AIKA is understood as 
the theory itself. The most faithful formalization of this view is the 
well-founded extension semantics but the stable-extension semantics, 
which extends Reiter's semantics to autoepistemic logic, coincides with 
the well-founded extension semantics wherever the autotheoremhood view 
seems to make sense. Thus, it was Reiter's default logic that for the 
first time incorporated into the reasoning process the principle of 
knowledge minimization, resulting in a better formalization of Moore's 
intuitions than Moore's own logic.

Fifteen years ago \citet{Halpern97} analyzed the intuitions of Reiter,
McDermott and Doyle, and Moore, and showed that there are alternative
ways, in which they could be formalized. Halpern's work suggested that
the logics proposed by Reiter, McDermott and Doyle, and Moore are not
necessarily ``determined'' by these intuitions. We argue here that by
looking more carefully at the informal semantics of those logics,
they do indeed seem ``predestined'' and can be derived in a systematic
and principled way from a few basic informal intuitions.

\ignore{Third, we noted that while the semantics of expansions proposed by
Moore is not a proper formalization of the autotheoremhood
perspective, it can be seen as a formalization of another dialect of
the autoepistemic language, suitable for modeling potential belief
states of a weaker sort of autoepistemic agent that is willing to rely
on or can be trapped in self-supported beliefs.}

\ignore{Finally, this paper also recalls the remarkable power of the fixpoint 
theory of nonmonotone operators of \citet{DeneckerMT00}, from which all 
major semantics of two of the most important nonmonotonic reasoning 
formalisms can be derived, and which, by its abstract nature, can be 
used to describe two opposite nonmonotonic reasoning: knowledge
minimization and knowledge maximization (through truth minimization).}

\section*{Acknowledgments}

The work of the first author was partially supported by FWO-Vlaanderen 
under project G.0489.10N. The work of the third author was partially 
supported by the NSF grant IIS-0913459.

{\small

}

\end{document}